\RequirePackage{fix-cm}
\documentclass{article}

\usepackage[T1]{fontenc}
\usepackage{arxiv}
\usepackage{authblk}

\usepackage{natbib}
\usepackage{amsmath,amssymb,amsthm}
\usepackage{thmtools,thm-restate}
\usepackage{graphicx}
\graphicspath{{numerical/}}
\usepackage{booktabs}
\usepackage{algorithmic}
\usepackage{algorithm}
\usepackage[normalem]{ulem}
\usepackage{xcolor}
\usepackage{bbm}
\usepackage{multirow}
\usepackage{enumitem}
\usepackage{colortbl}
\usepackage{tikz}
\usetikzlibrary{arrows.meta,decorations.pathreplacing}
\usepackage{float}
\usepackage[hidelinks]{hyperref}
\usepackage{url}
\usepackage{placeins}
\allowdisplaybreaks

\renewcommand{\headeright}{arXiv Preprint}
\renewcommand{\undertitle}{arXiv Preprint}
\renewcommand{\shorttitle}{Annealed L\'{e}vy Diffusion}

\newtheorem{theorem}{Theorem}[section]

\newtheorem{proposition}[theorem]{Proposition}

\newtheorem{remark}{Remark}[section]
\newtheorem{definition}{Definition}[section]

\newcommand{\R}{\mathbb{R}}
\newcommand{\E}{\mathbb{E}}

\newcommand{\bx}{\mathbf{x}}
\newcommand{\bn}{\mathbf{n}}
\newcommand{\bv}{\mathbf{v}}
\newcommand{\bz}{\mathbf{z}}
\newcommand{\by}{\mathbf{y}}
\newcommand{\btheta}{\boldsymbol{\theta}}
\newcommand{\eps}{\boldsymbol{\epsilon}}
\definecolor{aldblue}{RGB}{30,100,200}
\definecolor{aldred}{RGB}{200,60,40}
\definecolor{tblgray}{gray}{0.92}

\title{Annealed L\'{e}vy Diffusion via Gaussian Surrogate Reverse Dynamics}

\author[1]{Yuling Jiao}
\author[2]{Lican Kang}
\author[3]{Jerry Zhijian Yang}
\author[1]{Cheng Yuan}
\author[3]{Ruizhe Zhang}

\affil[1]{School of Artificial Intelligence, Wuhan University,
Wuhan, Hubei 430072, China. \protect\\
\texttt{\{yulingjiaomath,yuancheng\}@whu.edu.cn}}

\affil[2]{Institute for Math and AI, Hubei Key Laboratory of
Computational Science, \protect\\
and School of Artificial Intelligence, Wuhan University,
Wuhan, 430072, China. \protect\\
\texttt{kanglican@whu.edu.cn}}

\affil[3]{School of Mathematics and Statistics, Wuhan University,
Wuhan, Hubei 430072, China. \protect\\
\texttt{\{zjyang.math,ruizhezhang\}@whu.edu.cn}}

\date{October 14, 2026}

\begin{document}
\maketitle

\begin{abstract}
Gaussian diffusion models \citep{ho2020ddpm} offer a practical framework for generative modeling.
Denoising L\'{e}vy Probabilistic Models (DLPM) \citep{shariatian2025dlpm} introduce heavy-tailed perturbations with a fixed tail index throughout denoising.
We introduce \textbf{Annealed L\'{e}vy Diffusion (ALD)}.
ALD varies the stability index $\alpha(t)$ to combine heavy-tailed exploration with Gaussian refinement.
Our marginal-first formulation represents stable noise as a scalar Gaussian scale mixture.
This formulation enables standard Gaussian noise-prediction training with one additional scale input.
We study two sampling processes.
ALD-renoise estimates the clean sample and injects fresh marginal noise.
ALD-posterior preserves dependence between successive states through an accumulated conditional variance and a Gaussian bridge update.
Both samplers reuse the same trained predictor with different scalar inputs.
We evaluate ALD on four 2D benchmarks and an imbalanced nine-mode mixture.
We also examine training efficiency and the effects of prediction perturbations on clipped Holtsmark data.
Additional experiments evaluate MNIST and CelebA64 generation and CIFAR-10LT augmentation.
ALD preserves local geometry on the evaluated light-tailed and manifold-structured targets.
\end{abstract}

\section{Introduction}
\label{sec:intro}

Diffusion models \citep{sohldickstein2015deep,song2020score,ho2020ddpm} have established
themselves as the dominant paradigm for high-fidelity generative modeling,
with applications spanning image synthesis
\citep{dhariwal2021diffusion,rombach2022latent}
and scientific problems
\citep{pandey2024heavytailed,watson2023rfdiffusion,zeni2025mattergen}.
Their core mechanism is to corrupt data $\bx_0 \sim p_{\mathrm{data}}$
via a fixed stochastic process and learn to reverse it:
the forward process produces $\bx_t = m_t \bx_0 + \sigma_t \bn$ with
$\bn \sim \mathcal{N}(\mathbf{0}, \mathbf{I})$,
and generation proceeds via the estimated reverse kernel
$p_\theta(\bx_{t-1}\mid\bx_t)$.
The common choice of Gaussian noise is motivated by analytical tractability:
it admits closed-form forward kernels, a stable $\epsilon$-prediction objective,
and a reverse-time SDE representation under suitable regularity conditions
via Anderson's time-reversal formula \citep{anderson1982reverse}.
However, for target distributions with polynomial tails, the Gaussian terminal
prior and light-tailed perturbation kernels provide exponentially weaker relative
coverage of extreme regions than the data law.
With finite samples and training budgets, the network therefore struggles to estimate the score accurately in the tail regions
due to the scarcity of training samples there.

This score estimation difficulty is consequential for a broad and scientifically important class
of problems.
The Holtsmark distribution describes Coulomb force statistics from random point charges
and is a $3/2$-stable distribution
($\alpha = 3/2$, with PDF decaying as $|x|^{-5/2}$)
\citep{holtsmark1919broadening,samorodnitsky1994stable,nolan2020univariate};
financial return distributions can exhibit power-law tails, motivating
heavy-tailed alternatives to Gaussian models
\citep{gopikrishnan1999scaling,nolan2020univariate,larralde2025scaling};
heavy-tailed statistics also arise in fluid dynamics and related transport phenomena,
as illustrated by studies of anomalous transport and turbulence spreading
\citep{solomon1993observation,buonocore2020stochastic,milovanov2025turbulence}.
For such distributions, Gaussian diffusion can suffer from \emph{tail blindness}:
the terminal prior is light-tailed and the training distribution gives sparse
effective coverage to rare extremes, so learned scores in tail regions are
poorly constrained and may collapse toward bulk-dominated behavior.
This bias is not merely a representational limitation: increasing capacity alone
does not repair the sparse far-tail coverage induced by Gaussian corruption
\citep{song2019generative,yu2026heavy};
see also Experiment~\ref{exp:holtsmark_robustness}.

The natural remedy is to replace Gaussian noise with $\alpha$-stable L\'{e}vy noise
\citep{applebaum2009levy}, whose tail probability decays polynomially as
$\Pr(|z|>x)\sim C_\alpha x^{-\alpha}$ for $\alpha\in(0,2)$;
at the boundary $\alpha=2$ the distribution reduces to the Gaussian,
whose tails decay exponentially and the polynomial formula no longer applies.
This provides a parametric family that can be matched to infinite-variance
polynomial-tail targets with tail exponent in $(0,2)$ by choosing $\alpha$
appropriately.
However, prior attempts encounter one of two fundamental obstructions.
\emph{Continuous L\'{e}vy--It\^{o} SDE approaches} such as LIM
\citep{yoon2023lim} require a reverse-time formulation adapted to jump-driven dynamics.
Although L\'{e}vy processes are semimartingales and admit It\^{o} calculus,
the Brownian reverse-SDE formula based on the ordinary score
$\nabla_\bx \log p_t(\bx)$ cannot be applied directly.
For isotropic $\alpha$-stable noise with $\alpha<2$, the generator is nonlocal
and involves the fractional Laplacian $-(-\Delta)^{\alpha/2}$.
LIM derives the corresponding reverse-time formulation using a nonlocal
fractional score and learns it through fractional denoising score matching,
introducing additional analytical complexity relative to Gaussian diffusion.
\emph{Discrete Markov chain models} such as DLPM \citep{shariatian2025dlpm}
avoid this fractional-score formulation by representing stable noise as a
scalar Gaussian scale mixture and deriving approximate reverse updates through
conditional Gaussian posterior calculations.
Their explicit stable marginal laws rely on convolution closure among
independent $\alpha$-stable increments with a common stability index.
DLPM therefore keeps $\alpha$ fixed throughout the chain; varying the index
would generally take the accumulated noise outside the stable family used
in its derivation.
Although the noise amplitude can decrease during denoising, this fixed-index
construction does not support independently annealing the tail index toward
the Gaussian boundary $\alpha=2$ during final refinement.
This leaves no tail-index schedule for assigning different tail behavior
to exploration and geometry refinement.
Recent theoretical and empirical work also identifies a trade-off between
improved initialization and the statistical difficulty of reverse estimation
under heavy-tailed noising \citep{heavytradeoff2026}.
These paradigms expose complementary modeling constraints:
jump-driven formulations require nonlocal reverse operators, while standard
fixed-index stable chains do not provide a tail-index schedule.
This motivates constructions that keep conditional Gaussian calculations
tractable without requiring stable closure across time.
We introduce \textbf{Annealed L\'{e}vy Diffusion (ALD)} to address the
fixed-index restriction through a marginal-first construction that specifies
$\alpha(t)$-stable perturbation kernels directly at each time.
By retaining a scalar Gaussian scale-mixture representation, ALD allows
$\alpha(t)$ to vary while preserving tractable conditional Gaussian calculations
and standard Gaussian noise-prediction training with one additional scale input.
The resulting schedule combines heavy-tailed exploration with Gaussian refinement.

We adopt a marginal-first formulation related to diffusion models with
location-scale noise \citep{jolicoeur2023locationscale}.
Instead of deriving the forward process from discrete convolution or a L\'{e}vy SDE,
we define the perturbed state directly via the sub-Gaussian representation of
an $\alpha(t)$-stable vector:
\begin{equation*}
  \bx_t = m_t \bx_0 + \sigma_t \sqrt{S_t}\,\bn,
  \quad S_t \sim \mathcal{A}_{\alpha(t)/2},
  \quad \bn \sim \mathcal{N}(\mathbf{0},\mathbf{I}),
\end{equation*}
where $\mathcal{A}_{\alpha(t)/2}$ is a positive $(\alpha(t)/2)$-stable variance multiplier~\citep[Sec.~2.5, Prop.~2.5.2]
{samorodnitsky1994stable}.
Because $\bx_t$ is not derived from convolution of historical increments,
$\alpha(t)$ can be any prescribed function of $t$, enabling a two-phase
annealing schedule that transitions from heavy-tailed exploration
($t\geq\tau$) to pure Gaussian refinement
($t<\tau$, $S_t \equiv 1$).
The reverse process is a discrete ancestral sampler: conditioned on $S_t$, the
forward likelihood and denoising score target are Gaussian, so training reduces
to standard Gaussian score matching with a single additional scalar input
$\log S_t$.
The practical samplers reconstruct a clean-data estimate with a Gaussian
noise predictor. We study direct marginal re-noising (ALD-renoise) and a
variance-controlled adjacent transition (ALD-posterior) that retains the
current-state contribution and tracks the accumulated conditional noise variance.

\paragraph{Contributions.}
Our central contribution is a \emph{marginal-first} construction that uses
time-dependent $\alpha(t)$-stable perturbation kernels and conditional
Gaussian noise prediction.
This construction permits a two-phase schedule from heavy-tailed exploration
to Gaussian refinement and can reuse standard diffusion backbones with a
scalar scale-conditioning input.
We provide two practical reverse updates, ALD-renoise and ALD-posterior,
which share a trained predictor but differ in adjacent-state coupling.
Through experiments on synthetic and image datasets, we analyze tail recovery,
geometric fidelity, and sensitivity to prediction errors, and evaluate
the effectiveness of ALD for downstream data augmentation.

\section{Related Work}
\label{sec:related}

Score-based generative models \cite{song2020score,ho2020ddpm} learn to reverse
diffusion via score estimation; the score SDE framework \cite{song2021sde} unifies discrete and continuous
approaches, identifying the probability flow ODE as a deterministic counterpart to the
reverse SDE. 
Flow Matching \cite{lipman2022flow,liu2022flow} regresses directly on
conditional velocity fields, while Stochastic Interpolants
\cite{albergo2023stochastic,ma2024sit} subsume both perspectives
by deriving smooth marginal velocity fields from algebraic interpolation.
Among heavy-tailed generative models,
DLPM \cite{shariatian2025dlpm} replaces Gaussian noise with fixed-$\alpha$
stable noise in a discrete Markov chain via convolution closure;
LIM \cite{yoon2023lim} embeds $\alpha$-stable L\'{e}vy processes into a continuous
SDE and derives a reverse process via fractional denoising score matching;
t-EDM and t-Flow \cite{pandey2024heavytailed} use Student-t distributions
with a fixed Student-t tail parameter under the EDM framework.
DLPM and LIM use a fixed stable index, while t-EDM and t-Flow use a fixed
Student-t tail parameter. ALD prescribes a time-dependent stable index
through a conditional Gaussian scale-mixture formulation.
Cold Diffusion \cite{bansal2022cold} studies more general image degradations,
including deterministic transforms such as blur and masking, rather than
requiring stable-noise perturbations.
The sub-Gaussian representation $\mathbf{z} \stackrel{d}{=} \sqrt{S}\,\bn$ is classical
\cite{samorodnitsky1994stable,feller1971introduction} and has been exploited
in robust statistics and neural network weight analysis
\cite{hodgkinson2021multiplicative};
we use it as the structural foundation of our marginal-first forward process.
ALD's two-phase annealing draws philosophical inspiration from
Annealed Importance Sampling \cite{neal2001ais} and parallel tempering
\cite{swendsen1986replica}, but performs annealing within a single
continuous-time reverse trajectory via the dynamic $\alpha(t)$ parameter.

\section{Preliminaries}
\label{sec:background}

\subsection{Ancestral Sampling in the Gaussian Setting}
\label{sec:bg:gaussian}

Denoising diffusion probabilistic models (DDPMs) \cite{ho2020ddpm} specify a
discrete Markov forward chain
\begin{equation}
 q(\bx_{1:K}\mid\bx_0)=\prod_{k=1}^{K}q(\bx_k\mid\bx_{k-1}),
 \qquad
 q(\bx_k\mid\bx_{k-1})
 =\mathcal N(\sqrt{\alpha_k}\,\bx_{k-1},\,\beta_k\mathbf I),
 \quad \alpha_k=1-\beta_k.
 \label{eq:ddpm_markov_forward}
\end{equation}
Writing $\bar\alpha_k=\prod_{j=1}^{k}\alpha_j$, its one-time marginal is
\begin{equation}
 \bx_k=\sqrt{\bar\alpha_k}\,\bx_0
       +\sqrt{1-\bar\alpha_k}\,\bn,
 \qquad \bn\sim\mathcal N(\mathbf0,\mathbf I).
 \label{eq:ddpm_forward_marginal}
\end{equation}
The clean-data-conditioned adjacent-time posterior is
\begin{equation}
 q(\bx_{k-1}\mid\bx_k,\bx_0)
 =\mathcal N\!\left(A_k\bx_k+B_k\bx_0,\,
                    \tilde\beta_k\mathbf I\right),
 \label{eq:ddpm_adjacent_posterior}
\end{equation}
where
\[
 A_k=\frac{\sqrt{\alpha_k}(1-\bar\alpha_{k-1})}{1-\bar\alpha_k},
 \qquad
 B_k=\frac{\sqrt{\bar\alpha_{k-1}}\beta_k}{1-\bar\alpha_k},
 \qquad
 \tilde\beta_k=\frac{1-\bar\alpha_{k-1}}{1-\bar\alpha_k}\beta_k.
\]
Standard ancestral DDPM sampling parameterizes this adjacent-time Gaussian
reverse kernel and substitutes a network estimate $\hat\bx_0(\bx_k,k)$ for
$\bx_0$ in its mean. Thus the update retains an explicit $\bx_k$ term and
uses the adjacent-posterior variance $\tilde\beta_k$; it is not an independent
draw from the next marginal
$\mathcal N(\sqrt{\bar\alpha_{k-1}}\hat\bx_0,
            (1-\bar\alpha_{k-1})\mathbf I)$.
The usual $\epsilon$-prediction parameterization trains the network by
minimizing $\E\|\eps_{\btheta}(\bx_k,k)-\bn\|_2^2$.

\subsection{Heavy-Tailed Diffusion: DLPM}
\label{sec:bg:dlpm}

DLPM \cite{shariatian2025dlpm} extends the above by replacing the Gaussian noise
$\bn$ with $\alpha$-stable noise. A non-degenerate random variable $X$ is \emph{$\alpha$-stable},
with stability index $\alpha \in (0,2]$, if, for every integer $n \geq 2$,
independent random variables $X_1,\ldots,X_n$ having the same distribution as
$X$ satisfy
$X_1+\cdots+X_n \stackrel{d}{=} n^{1/\alpha}X+d_n$
for some constant $d_n \in \mathbb{R}$.
For $\alpha \in (0,2)$, the distribution is heavy-tailed with polynomial tail
$\Pr(|X|>x) \sim C_\alpha x^{-\alpha}$, where the stability index $\alpha$
directly controls the polynomial decay rate.
The boundary case $\alpha = 2$ is Gaussian, whose tails decay exponentially---it
does not satisfy the polynomial formula above.

DLPM defines the forward process as a discrete Markov chain with transition
\begin{equation}
\bx^{(k)} \mid \bx^{(k-1)},S_k
\sim
\mathcal{N}\!\left(
\gamma_k\bx^{(k-1)},
\sigma_k^2 S_k\mathbf{I}
\right),
\qquad
S_k \sim \mathcal{A}_{\alpha/2}.
\label{eq:dlpm_forward}
\end{equation}
Here, $S_k$ is an independently sampled positive $(\alpha/2)$-stable random variable at each step; specifically, for $0<\alpha<2$, it is supported on $(0,\infty)$ and characterized by the Laplace transform $\mathbb{E}\!\left[e^{-\lambda S_k}\right]=\exp\!\left(-\lambda^{\alpha/2}\right)$ for $\lambda\geq 0$, exhibiting increasingly heavy-tailed behavior as $\alpha$ decreases.
By convolution closure (the sum of two independent $\alpha$-stable
variables with the \emph{same} $\alpha$ is again $\alpha$-stable),
the marginals admit the closed-form Gaussian scale-mixture representation
\begin{equation*}
\bx^{(k)} \mid \bx_0
\stackrel{d}{=}
\bar{m}_k\bx_0
+\bar{\sigma}_k\sqrt{\bar{S}_k}\,\bn,
\qquad
\bar{S}_k\sim\mathcal{A}_{\alpha/2},
\qquad
\bn\sim\mathcal{N}(\mathbf{0},\mathbf{I}),
\qquad
\bar{S}_k\perp\bn.
\end{equation*}
Here $\gamma_k$ and $\sigma_k$ are one-step coefficients, whereas
$\bar m_k$ and $\bar\sigma_k$ are the marginal signal and noise scales.

This yields a training objective analogous to $\epsilon$-prediction at comparable
cost to DDPM. Fixed-index heavy-tailed perturbations can introduce
spurious extremes near geometric structures, as illustrated by our evaluated
2D benchmarks (Section~\ref{exp:synthetic}).
If independent stable increments have different indices, their sum is
generally not stable, so the fixed-index marginal representation above
does not directly extend to a time-varying index. A different construction
is therefore needed to prescribe stable perturbations with a varying index.
This modeling constraint motivates our marginal-first formulation.

\subsection{Notation}
\label{sec:bg:notation}

Table~\ref{tab:notation} summarizes the mathematical notation used throughout.

\begin{table}[htbp]
\centering
\caption{Summary of Notation}
\label{tab:notation}
\resizebox{\linewidth}{!}{\begin{tabular}{ll}
\toprule
\textbf{Symbol} & \textbf{Description} \\
\midrule
$\bx_t$ & State vector at diffusion time $t \in [0, 1]$ \\
$\bx_0$ & Data sample from $p_{\mathrm{data}}(\bx)$ \\
$\alpha(t)$ & Time-dependent stability index, $\alpha(t) \in (1, 2]$ \\
$S_t$ & Positive stable variance multiplier, $S_t\sim\mathcal A_{\alpha(t)/2}$; $S_t\equiv1$ if $\alpha(t)=2$ \\
$\tilde{S}_t$ & Clamped impulse, $\tilde{S}_t = \min(S_t, S_{\max})$ \\
$\bn$ & Standard Gaussian random vector, $\bn \sim \mathcal{N}(\mathbf{0}, \mathbf{I})$ \\
$\btheta$ & Learnable parameters of the neural network \\
$m_t, \sigma_t$ & Signal scale and noise scale coefficients \\
$\tau$ & Phase-transition threshold (refinement phase: $t < \tau$) \\
$W_2(p,q)$ & Wasserstein-2 distance; experimental columns use the sliced version (Section~\ref{sec:exp:setup}) \\
\bottomrule
\end{tabular}}
\end{table}

\section{Methodology}
\label{sec:method}
\raggedbottom

\subsection{ALD Forward Process and Annealing Schedule}
\label{sec:method:forward}

Instead of relying on discrete Markov chains or Brownian-driven SDEs,
we define the forward process directly through time-dependent marginal
perturbation kernels.
We employ the \emph{sub-Gaussian representation} of $\alpha$-stable
vectors \citep{samorodnitsky1994stable}: the noise vector can be written
as $\bz_t=\sqrt{S_t}\,\bn$, where $\bx_0$, $S_t$, and $\bn$ are mutually independent.
At $\alpha(t)=2$, this representation degenerates to $S_t\equiv1$.

With stability index $\alpha(t)\in(0,2]$, the characteristic function
of $\bz_t$ is
\begin{equation}
\mathbb E\!\left[e^{i\langle\boldsymbol{\xi},\bz_t\rangle}\right]
=
\exp\!\left[
-\left(\frac{\|\boldsymbol{\xi}\|_2^2}{2}\right)^{\alpha(t)/2}
\right].
\label{eq:stable_gaussian_normalized_cf}
\end{equation}
At the boundary $\alpha(t)=2$, this distribution recovers the standard
Gaussian $\mathcal N(\mathbf0,\mathbf I)$.
Using the representation mentioned in Section \ref{sec:intro}, we define the ALD forward process
as follows.

\begin{definition}[ALD Forward Process]
\label{def:ald_forward}
At any diffusion time $t\in[0,1]$, the perturbed state $\bx_t$ is
defined via the sub-Gaussian representation of an isotropic
$\alpha(t)$-stable vector:
\begin{equation}
    \bx_t=m_t\bx_0+\sigma_t\sqrt{S_t}\,\bn,
  \qquad
  \bn\sim\mathcal N(\mathbf0,\mathbf I),
  \qquad
  S_t\sim\mathcal A_{\alpha(t)/2},
  \label{eq:marginal_forward}
\end{equation}
where $\bx_0\sim p_{\mathrm{data}}$,
$m_t,\sigma_t$ are defined by Eq.~\eqref{eq:trig_schedule},
and $S_t$ is sampled independently of $\bn$.
For $\alpha(t)=2$, we set $S_t\equiv1$.
\end{definition}

For the interpolation coefficients $m_t$ and $\sigma_t$, we choose
the trigonometric schedule~\citep{chen2023importance}:
letting $\vartheta_t=\frac{\pi}{2}t$, we set
\begin{equation}
m_t=\cos(\vartheta_t),
\qquad
\sigma_t=\sin(\vartheta_t).
\label{eq:trig_schedule}
\end{equation}

The forward perturbation can be interpreted as a direct perturbation
of $\bx_0$ at time $t$: $m_t$ attenuates the data signal,
$\sigma_t$ scales the noise level, and $\alpha(t)$ governs the tail
behavior.
When $\alpha(t)<2$, the noise exhibits a heavy polynomial tail,
$\Pr(\|\bz_t\|_2>r)\asymp r^{-\alpha(t)}$ as $r\to\infty$,
where the stability index dictates the polynomial decay rate.
At $\alpha(t)=2$, it transitions to a Gaussian distribution with
exponential tail decay.
Thus, the construction generalizes the one-time DDPM perturbation
marginals (Section~\ref{sec:bg:gaussian}) to heavy-tailed noise.
Setting $\alpha(t)\equiv2$ recovers the DDPM Gaussian marginal kernel
$q(\bx_t\mid\bx_0)$.

Crucially, because the construction directly defines the
\emph{marginal} at each $t$ rather than accumulating convolution steps,
$\alpha(t)$ can be prescribed independently across arbitrary time steps.
This flexibility enables a time-varying schedule for $\alpha(t)$ in practice.

\begin{remark}[Comparison with SDE]
Classical score-SDE constructions rely on a continuous-time forward dynamics
whose terminal law is chosen or approximated by an easy-to-sample prior, and whose
reverse-time dynamics are justified by Anderson's
time-reversal formula.
ALD does not make this SDE commitment.
Instead, \eqref{eq:marginal_forward} defines a family of marginal perturbation
kernels directly. The terminal marginal is exactly sampleable because
$m_1=0$ and $\sigma_1=1$, giving
$\bx_1=\sqrt{S_1}\bn$, independent of $\bx_0$.
\end{remark}

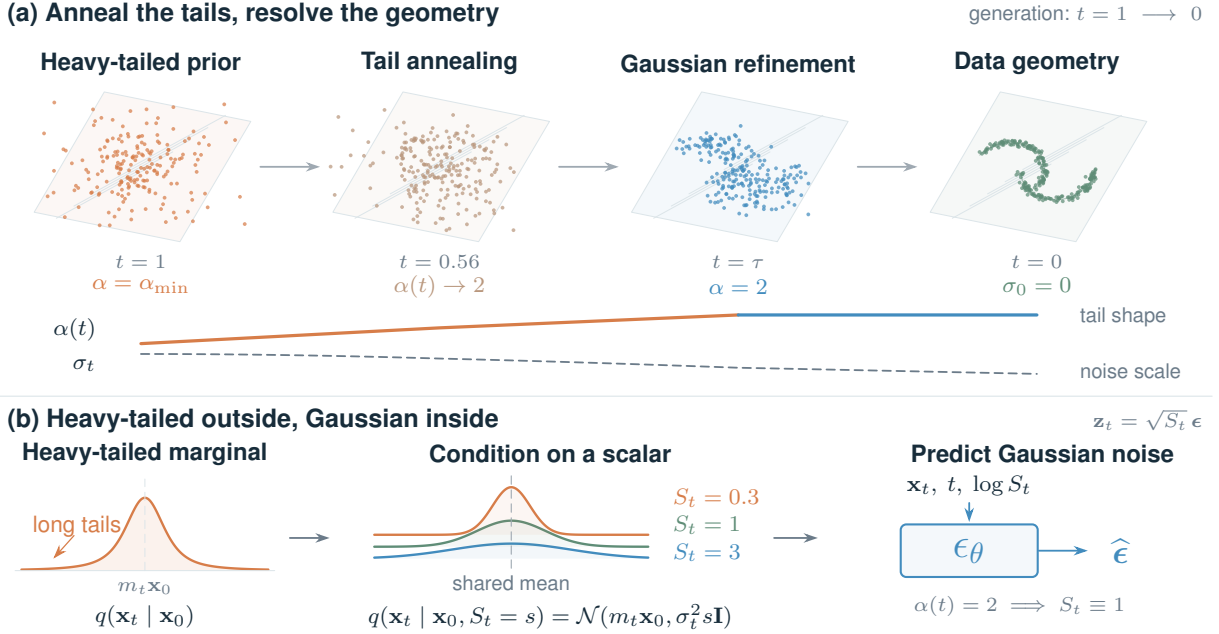
\begin{figure}[t]
\centering

\begingroup
\definecolor{ALDInk}{HTML}{172C3A}
\definecolor{ALDMuted}{HTML}{6B7B89}
\definecolor{ALDOrange}{HTML}{D77D49}
\definecolor{ALDSand}{HTML}{B9977D}
\definecolor{ALDBlue}{HTML}{428CBE}
\definecolor{ALDGreen}{HTML}{5F8D77}
\definecolor{ALDGrid}{HTML}{DCE5E9}
\resizebox{\linewidth}{!}{%
\begin{tikzpicture}[
  x=1cm,y=1cm,
  font=\sffamily\fontsize{8}{9.5}\selectfont,
  text=ALDInk,line cap=round,line join=round,
  ald arrow/.style={-{Stealth[length=1.8mm,width=1.2mm]},
                    draw=ALDMuted,line width=.65pt},
  ald heading/.style={anchor=west,font=\sffamily\bfseries\fontsize{9}{10.5}\selectfont},
  ald stage/.style={font=\sffamily\bfseries\fontsize{8.5}{10}\selectfont},
  ald small/.style={font=\sffamily\fontsize{7.2}{8.5}\selectfont,text=ALDMuted}
]
\path[use as bounding box] (-.12,-.06) rectangle (15.55,8.12);

\node[ald heading] at (0,7.86)
  {(a) Anneal the tails, resolve the geometry};
\node[ald small,anchor=east] at (15.25,7.86)
  {generation: $t=1\;\longrightarrow\;0$};
\node[ald stage] at (1.80,7.25) {Heavy-tailed prior};
\node[ald stage] at (5.55,7.25) {Tail annealing};
\node[ald stage] at (9.30,7.25) {Gaussian refinement};
\node[ald stage] at (13.05,7.25) {Data geometry};

\begin{scope}[shift={(1.80,5.95 )}]
\path[fill=ALDOrange,fill opacity=.055,draw=ALDGrid,line width=.35pt]
  (-1.34,-.57)--(.49,-.94)--(1.38,.57)--(-.46,.94)--cycle;
\draw[ALDGrid,line width=.25pt] (-1.0620,-0.6455) -- (0.1780,0.0345);
\draw[ALDGrid,line width=.25pt] (-0.8410,-0.4928) -- (0.3990,0.1873);
\draw[ALDGrid,line width=.25pt] (-0.6200,-0.3400) -- (0.6200,0.3400);
\draw[ALDGrid,line width=.25pt] (-0.3990,-0.1873) -- (0.8410,0.4928);
\draw[ALDGrid,line width=.25pt] (-0.1780,-0.0345) -- (1.0620,0.6455);
\foreach \aldpx/\aldpy in {
-0.1653/-0.1273, 0.1135/-0.0116, 0.2459/0.1173, -0.0466/0.2600, 0.1383/0.0357, -0.8348/-0.2950, -0.1916/0.1878, 0.0409/-0.0241, 0.0305/-0.2536,
-0.1069/0.1188, -0.1232/-0.1764, 0.7547/0.1502, -0.3549/0.5898, 0.3807/-0.2121, 0.0459/-0.1758, -0.2574/-0.0345, -1.0916/-0.0501, -0.1488/-0.0449,
-0.9733/0.2200, -0.4212/0.1286, 0.1192/0.1639, 0.6768/0.1635, 0.5896/-0.5633, -0.3036/-0.0133, -0.2756/-0.5333, -0.1557/-0.3413, 0.2307/0.2987,
-0.1638/0.4133, -0.1482/-0.2453, 0.2055/0.0242, 0.0723/0.5460, 0.6313/-0.3253, 0.2779/-0.4696, 0.0796/0.2113, 0.6838/0.0229, 0.7290/0.2938,
-0.6872/-0.0367, -0.3001/0.1487, -0.7817/-0.5068, -0.2393/0.3163, 0.8047/-0.3365, 0.7917/0.3750, -0.0366/-0.0669, -0.5129/0.2170, -0.2604/0.0309,
-1.0964/-0.3243, -0.6254/-0.3280, -0.1562/-0.2638, -0.8160/-0.2377, -0.2772/0.3476, 0.6998/-0.0637, -0.2364/-0.3659, -0.0287/0.1749, 0.3318/0.2770,
-0.2679/-0.2251, -0.4236/0.1123, -0.5300/-0.3518, 0.2137/0.6123, -0.2215/-0.4408, -0.4545/0.2477, -0.0794/-0.1961, 0.4069/0.4757, -0.1304/-0.1428,
-0.3023/0.0105, 1.1820/0.3358, 0.0705/0.7219, -0.3436/0.1375, -0.3611/-0.3168, -0.0933/0.1329, -0.2272/0.5081, -0.4964/0.6149, -0.3308/-0.6256,
-0.3529/0.3572, -0.2512/-0.0226, -0.1008/0.1143, -0.4295/-0.1861, 0.6529/-0.6838, -0.2168/0.0730, 0.4126/0.0498, -0.2894/0.0520, 0.0594/-0.5307,
-0.5421/-0.0932, 0.3130/-0.6939, 0.4377/-0.6428, 0.4077/0.7492, -0.0569/0.1308, -0.0162/0.3816, 0.1916/0.5472, -0.1999/-0.4009, 0.5128/0.1108,
-0.3912/-0.7345, -0.0462/-0.1595, 0.1035/0.1806, -0.4041/-0.5745, -0.0803/-0.1668, -0.6614/-0.0685, -0.0703/0.0826, 0.8538/0.2739, 0.7142/0.0899,
0.5281/0.1543, 0.0284/-0.1715, 0.0015/-0.0129, -0.2536/0.3917, -0.0338/0.3739, 0.2410/0.0136, -0.1818/0.1465, -0.2924/0.1924, -0.5180/-0.7344,
0.4526/-0.2100, -0.5213/-0.3344, -0.4752/0.0673, 0.2034/-0.0701, -0.2372/0.0378, 0.0129/0.1553, -0.0437/0.2332, -0.1447/-0.1556, -0.8669/0.0439,
-1.2163/0.4195, 0.8382/-0.5088, -0.1317/0.7002, -0.2851/-0.0122, -0.0477/-0.0629, 0.2971/-0.3882, -0.7345/0.0821, -0.3234/0.1853, -0.4909/0.1284,
-0.6152/-0.3203, 0.6605/0.4446, 0.0745/-0.0699, -0.9554/-0.6317, -1.1712/-0.3707, 0.1720/-0.0456, -0.0288/-0.2419, -0.4182/0.3216, 0.3326/0.1605,
-0.2869/-0.3433, 0.3123/-0.0957, -0.5493/-0.2559, -0.4883/-0.0518, -0.2428/-0.1913, 0.4302/0.0919, -0.9429/0.2036, 0.0480/-0.2594, 0.2045/-0.0927,
0.0800/-0.0030, -0.2384/-0.2874, -0.2287/0.1661, -0.1133/-0.5516, -0.2479/-0.1795, -1.0876/0.8074, 0.5519/0.3226, 0.1051/-0.3153, 0.0164/0.3883,
1.3033/-0.7187, 0.4124/-0.5739, -0.0067/-0.1119, 0.0268/0.1419, 0.0784/-0.1710, 0.2872/-0.1226, 0.1665/-0.0495, 0.3942/0.0053, -0.2484/-0.1978,
-0.6015/0.7206, 1.0464/0.7843, -0.6550/-0.1557, -0.4233/0.3327, 0.3436/0.2118, 0.6043/0.3562, 1.2435/-0.3768, 0.0106/0.3053, -0.1567/0.3795,
-0.1252/0.0256, 0.2115/0.3191, -0.7871/-0.2657, 0.3475/0.2129, 0.2065/0.3994, -0.1212/-0.6612, 0.3767/0.1023, 0.2527/-0.5305, 1.0682/0.5576,
-0.5912/0.5225, 0.1065/0.0887, 0.1819/-0.3335, -0.1701/-0.1401, -0.0442/-0.2350, 0.0460/0.2841, 0.3346/-0.0582, 0.4860/0.0982, 0.1347/0.1957,
-0.1626/-0.2145, -0.5753/0.5736, 0.1072/-0.2875, 0.6445/-0.4150, -0.2739/-0.0817
} {
  \fill[ALDOrange,opacity=.72] (\aldpx,\aldpy) circle[radius=.024];
}
\end{scope}
\node[ald small] at (1.80,4.76) {$t=1$};
\node[text=ALDOrange] at (1.80,4.45) {$\alpha=\alpha_{\min}$};

\begin{scope}[shift={(5.55,5.95 )}]
\path[fill=ALDSand,fill opacity=.055,draw=ALDGrid,line width=.35pt]
  (-1.34,-.57)--(.49,-.94)--(1.38,.57)--(-.46,.94)--cycle;
\draw[ALDGrid,line width=.25pt] (-1.0620,-0.6455) -- (0.1780,0.0345);
\draw[ALDGrid,line width=.25pt] (-0.8410,-0.4928) -- (0.3990,0.1873);
\draw[ALDGrid,line width=.25pt] (-0.6200,-0.3400) -- (0.6200,0.3400);
\draw[ALDGrid,line width=.25pt] (-0.3990,-0.1873) -- (0.8410,0.4928);
\draw[ALDGrid,line width=.25pt] (-0.1780,-0.0345) -- (1.0620,0.6455);
\foreach \aldpx/\aldpy in {
-0.0131/-0.1311, 0.5076/-0.1794, 0.1274/0.2471, 0.0865/-0.0451, -0.1282/0.1878, 0.7116/-0.0352, -0.4375/-0.1638, 0.1418/0.0430, 0.1183/-0.0591,
0.0383/-0.1956, -0.3167/0.3425, 0.3011/-0.2756, 0.8372/0.2800, -0.0352/0.2428, 0.2342/-0.2480, 0.2083/0.4198, -0.4803/0.1463, -0.1199/-0.1158,
-0.1897/-0.1940, -0.3667/0.2357, -0.3815/-0.0567, 0.3475/-0.0235, 0.0247/0.2391, -0.4486/0.5562, 0.6810/-0.5453, 0.0726/-0.1150, 0.9278/-0.7955,
0.0507/-0.5204, -0.2875/-0.4532, 0.6275/0.0991, -0.3532/0.3373, 0.2886/-0.3321, 0.1505/-0.1004, 0.4466/0.1032, 0.4703/-0.3159, 0.4498/-0.6150,
0.1699/0.6507, -0.2111/0.3055, 0.2551/-0.0497, -1.3680/0.0023, -0.4149/0.2139, 0.2429/-0.2441, -0.1991/0.3188, 0.7633/-0.1861, 0.3367/0.4328,
0.5599/0.2142, -0.3206/0.1281, -0.4711/-0.5054, -0.0878/-0.0959, -0.3049/-0.1274, -0.9496/0.1381, -0.2983/-0.2776, -0.1955/-0.1272, -0.3332/-0.0701,
-0.8366/0.0441, -0.5409/0.4908, 0.6922/-0.0682, -0.1912/0.0255, 0.0216/0.0690, 0.4185/0.3903, -0.3596/0.1967, -0.3283/-0.6605, -0.3088/0.1345,
-0.4294/0.2357, -0.3221/-0.3085, 0.1300/0.2462, -0.0139/-0.4888, 0.1175/0.3665, -0.4827/0.0339, 0.3689/-0.2451, 0.5842/0.1071, -0.1953/-0.4317,
-0.3534/-0.2332, 0.7248/0.0316, 0.0877/0.4453, -1.2708/0.6259, 0.2257/0.0160, -0.2745/-0.6971, -0.0307/0.0113, -0.0187/0.1470, -0.2092/-0.1042,
0.0670/0.1558, 0.0858/-0.7961, -0.6915/0.7141, 0.2107/-0.0499, -0.1040/0.2214, -0.8702/-0.2689, 0.5000/-0.5873, -0.0501/-0.0021, 0.8885/-0.0160,
-0.1459/-0.1918, 0.0412/-0.6523, -1.2303/-0.0005, -0.4478/-0.0271, 0.6868/-0.5849, 0.2332/-0.5934, 0.6785/0.3158, -0.1476/-0.0512, 0.4360/0.2477,
0.0163/0.0263, -0.1604/-0.3517, 0.2798/0.0390, 0.7263/0.3764, -0.2752/-0.5313, -0.1994/0.0327, -0.2200/0.3212, 0.0097/-0.6488, 0.0192/-0.2880,
-0.3909/-0.2939, -1.3169/-0.3527, -0.3068/0.3334, 0.4399/0.0496, 0.5056/0.2544, 0.2966/0.3278, 0.2375/-0.4617, -0.4577/0.1064, -0.1463/0.3402,
-0.8950/0.3100, -0.0877/0.4871, 0.9330/-0.0666, -0.4663/0.1731, -0.4205/0.3060, 0.0061/-0.4228, -0.1147/-0.0054, 0.2555/-0.1021, 0.1742/-0.0192,
-0.3674/-0.3746, -0.7052/0.1808, 0.5143/-0.2143, -0.4390/0.2267, 0.3368/-0.1260, 0.3433/-0.7223, -0.4418/0.3051, -0.1693/0.0272, -0.4715/-0.1151,
-0.0575/-0.1387, 0.7903/-0.5133, -0.2099/0.3829, -0.0264/-0.1371, 0.1013/-0.3552, -0.3270/0.3238, 0.2145/-0.3500, 0.2346/0.3508, -0.4683/0.1028,
-0.3565/-0.0032, -1.0902/0.4574, -0.1087/0.4301, -0.2721/-0.0495, -0.0265/0.3852, -0.0656/-0.1269, 0.6496/0.1851, -0.2425/-0.5708, -0.7883/-0.4892,
0.5707/0.2054, -0.1183/0.1533, -0.1862/-0.2129, -0.1406/0.0242, 0.3532/0.5126, -0.2386/-0.3931, -0.1249/0.2405, 0.4558/-0.1891, -0.3170/-0.2688,
-0.4748/0.1390, -0.2047/-0.2959, 0.2549/0.1978, -1.1162/0.1442, 0.4651/-0.3992, 0.6553/-0.1031, 0.0236/-0.1475, 0.0751/-0.3904, 0.5849/0.2299,
-0.2286/0.0761, 0.3385/-0.5389, 0.1989/-0.2378, 0.0882/0.1583, 0.2255/0.4165, 0.1243/-0.7768, 0.0403/0.2561, 0.0306/0.0027, 0.0462/-0.1181,
0.0929/0.1177, 0.6334/-0.4500, 0.5827/-0.3117, 0.3056/0.3016, 0.1806/0.0097, 0.2296/-0.1059, -0.2892/-0.3153, 0.1252/-0.0348, 0.4176/0.2782,
-0.3822/-0.3398, 0.4356/0.2803, -0.2969/0.3551, 0.4989/-0.0471, 0.4041/0.0813, 0.4530/-0.2799, -0.4288/0.4147, -0.3171/0.3581, -0.0775/0.0953,
0.1334/0.3101, -0.8689/0.0153, -0.0675/0.3422, 0.3545/0.0884, -0.0492/-0.7065, 0.1168/0.0053, -0.2324/0.3328, -0.1495/-0.0766, 0.1904/0.3880,
-0.4739/0.2415, -0.2952/-0.2349, 0.1593/-0.1844, 0.5817/-0.4325, -0.2795/-0.6186, -0.1777/-0.1312, 0.3902/-0.3450, 0.3635/-0.0312, 0.1586/0.1085,
0.2663/-0.2258, 0.5443/0.1001, -0.4427/-0.0411, 0.1012/-0.2873, 0.6196/-0.2908, -0.5358/0.1149
} {
  \fill[ALDSand,opacity=.72] (\aldpx,\aldpy) circle[radius=.024];
}
\end{scope}
\node[ald small] at (5.55,4.76) {$t=0.56$};
\node[text=ALDSand] at (5.55,4.45) {$\alpha(t)\rightarrow2$};

\begin{scope}[shift={(9.30,5.95 )}]
\path[fill=ALDBlue,fill opacity=.055,draw=ALDGrid,line width=.35pt]
  (-1.34,-.57)--(.49,-.94)--(1.38,.57)--(-.46,.94)--cycle;
\draw[ALDGrid,line width=.25pt] (-1.0620,-0.6455) -- (0.1780,0.0345);
\draw[ALDGrid,line width=.25pt] (-0.8410,-0.4928) -- (0.3990,0.1873);
\draw[ALDGrid,line width=.25pt] (-0.6200,-0.3400) -- (0.6200,0.3400);
\draw[ALDGrid,line width=.25pt] (-0.3990,-0.1873) -- (0.8410,0.4928);
\draw[ALDGrid,line width=.25pt] (-0.1780,-0.0345) -- (1.0620,0.6455);
\foreach \aldpx/\aldpy in {
0.0777/-0.1207, 0.6135/-0.2530, 0.0068/0.2813, 0.0631/-0.2378, -0.1202/-0.1164, 0.7016/-0.1456, -0.2548/-0.4419, -0.3295/-0.1302, 0.4734/-0.1925,
0.1336/-0.0628, 0.0228/-0.0071, -0.3531/0.3784, 0.5843/-0.2179, 0.4139/0.2515, 0.2620/-0.1605, -0.4650/0.0837, 0.1314/0.2546, 0.0933/-0.4841,
-0.4858/0.2492, 0.3858/-0.1469, -0.1882/-0.2608, -0.2353/0.2812, -0.2579/-0.1797, 0.3195/-0.3088, -0.4019/0.2508, -0.6091/0.3518, 0.6316/-0.4463,
0.4933/-0.1547, 0.8796/-0.5976, 0.2185/-0.5000, -0.2810/-0.3519, 0.7392/-0.1072, -0.4715/0.3632, 0.5631/-0.2761, 0.1015/-0.1643, 0.6283/-0.1339,
0.2641/-0.2458, 0.2192/-0.1540, 0.0130/0.3326, -0.3741/0.3342, 0.8817/-0.2101, 0.0431/-0.2101, -0.1525/0.1042, -0.5098/0.2650, 0.5808/-0.2226,
-0.1643/0.3003, 0.1674/0.1286, 0.5793/-0.1085, 0.3291/0.0798, -0.4339/0.2724, -0.3490/-0.4380, 0.2770/-0.3156, -0.2468/-0.2147, -0.2796/0.1919,
0.3498/-0.1790, -0.0092/-0.0416, -0.3468/0.1493, -0.5409/0.2718, -0.1648/-0.0729, 0.3751/-0.0419, -0.1952/0.1773, 0.0577/-0.0521, 0.1678/0.2012,
0.3066/0.0231, 0.1010/-0.6077, -0.3781/0.2701, -0.4170/0.2923, -0.0645/-0.1841, 0.0292/-0.1048, 0.1048/-0.4796, 0.0535/0.0099, -0.1904/-0.2381,
0.6142/-0.2079, 0.7319/-0.0035, -0.0683/-0.4000, -0.0261/-0.3653, 0.1563/-0.2157, 0.0455/-0.2099, -0.3571/0.3179, 0.1339/-0.2384, 0.3421/-0.3807,
0.2059/-0.3415, 0.1595/-0.1998, -0.0253/-0.1598, 0.3669/-0.0984, 0.4747/-0.5321, -0.3988/0.4246, 0.5762/-0.0509, -0.0663/0.2295, -0.1612/0.0915,
0.4198/-0.5346, 0.0635/-0.0496, 0.8456/-0.0818, 0.2266/-0.3660, -0.0422/-0.0474, -0.3976/-0.1979, -0.1611/0.0468, 0.7916/-0.3560, 0.0743/-0.4834,
0.7635/0.0140, -0.1837/-0.1603, 0.6665/-0.0063, -0.0768/-0.2495, -0.1492/-0.3440, 0.1725/-0.1017, -0.2535/0.2726, -0.2137/-0.4229, -0.2502/0.2125,
-0.4061/0.3436, 0.3212/-0.5321, 0.0517/-0.3818, -0.1140/-0.3926, -0.3844/-0.3625, -0.3344/0.3526, 0.0303/0.5135, 0.1012/-0.1041, -0.3460/0.2410,
-0.1686/0.3106, 0.3229/-0.4982, -0.6842/0.1702, -0.1137/0.3460, -0.7361/0.3356, -0.0484/-0.1853, 0.7879/-0.1334, -0.2679/-0.0866, -0.4449/0.3365,
0.4214/-0.3234, -0.0570/0.1536, 0.1112/0.0855, -0.1601/0.1662, -0.0802/-0.2588, -0.7733/0.2303, 0.6229/-0.2699, -0.2193/0.2689, 0.4918/-0.3171,
0.5579/-0.5054, -0.6223/0.2583, -0.1256/0.1772, -0.3144/-0.1915, 0.2164/-0.3112, 0.7059/-0.4786, -0.2825/0.4084, 0.5418/-0.1798, 0.2088/-0.4540,
-0.0630/0.1934, 0.1442/-0.2924, -0.0277/0.3835, -0.1373/0.0907, -0.0762/-0.2659, -0.3497/0.3486, 0.0372/0.1578, -0.0691/0.1015, -0.4153/0.3218,
-0.1642/-0.1272, 0.0435/-0.0072, 0.2400/-0.4538, -0.2910/-0.4498, 0.5612/-0.1550, -0.3446/0.2733, -0.2256/0.1200, -0.0300/-0.1023, -0.3301/0.3512,
-0.1491/-0.3389, -0.1173/-0.0860, 0.3080/-0.1682, -0.2019/-0.2748, -0.3363/0.2468, -0.0538/-0.2431, 0.0986/0.2349, 0.2325/-0.1946, 0.6439/-0.3250,
0.7822/-0.0652, 0.0018/-0.2189, 0.2337/-0.4358, 0.6552/-0.1003, -0.1615/-0.0775, -0.0676/-0.0851, 0.6011/-0.3363, 0.5697/-0.1572, 0.4743/-0.0183,
-0.0434/0.3993, -0.1189/-0.2456, 0.0428/-0.0277, 0.1431/-0.4064, -0.0609/0.1523, 0.0774/-0.0331, 0.1084/0.0160, 0.7265/-0.1949, 0.6059/-0.3527,
0.1975/0.1617, 0.0958/0.0660, 0.1503/-0.1606, -0.2470/-0.3235, 0.3351/-0.2296, 0.2550/0.1391, 0.4229/-0.2709, 0.2726/0.0773, -0.2183/0.3533,
0.3527/-0.3116, 0.2422/-0.0913, 0.2676/-0.2939, -0.6525/0.2533, -0.2010/-0.1244, 0.0037/0.1177, -0.0935/0.0215, -0.7023/0.2235, -0.3966/0.3294,
0.3632/-0.1886, 0.0122/-0.5869, -0.0744/-0.0597, -0.0499/0.1368, -0.3650/0.1836, -0.1187/0.3686, -0.3663/0.0592, -0.6910/0.3118, -0.1090/-0.1615,
0.1491/-0.3489, 0.6828/-0.3060, -0.1776/-0.5165, -0.0673/-0.0327, 0.6118/-0.3576, 0.5079/-0.2589, -0.0844/0.2177, 0.2421/-0.3683, 0.7133/0.0057,
-0.4825/0.1734, -0.1818/0.3426, -0.0556/0.1254, 0.2715/-0.0134, -0.6111/0.2277
} {
  \fill[ALDBlue,opacity=.72] (\aldpx,\aldpy) circle[radius=.024];
}
\end{scope}
\node[ald small] at (9.30,4.76) {$t=\tau$};
\node[text=ALDBlue] at (9.30,4.45) {$\alpha=2$};

\begin{scope}[shift={(13.05,5.95 )}]
\path[fill=ALDGreen,fill opacity=.055,draw=ALDGrid,line width=.35pt]
  (-1.34,-.57)--(.49,-.94)--(1.38,.57)--(-.46,.94)--cycle;
\draw[ALDGrid,line width=.25pt] (-1.0620,-0.6455) -- (0.1780,0.0345);
\draw[ALDGrid,line width=.25pt] (-0.8410,-0.4928) -- (0.3990,0.1873);
\draw[ALDGrid,line width=.25pt] (-0.6200,-0.3400) -- (0.6200,0.3400);
\draw[ALDGrid,line width=.25pt] (-0.3990,-0.1873) -- (0.8410,0.4928);
\draw[ALDGrid,line width=.25pt] (-0.1780,-0.0345) -- (1.0620,0.6455);
\foreach \aldpx/\aldpy in {
0.1200/-0.0974, 0.5776/-0.2591, -0.0757/0.2562, 0.0272/-0.3511, -0.0967/-0.2881, 0.6817/-0.1675, -0.1873/-0.2679, -0.1713/-0.0750, 0.6036/-0.3061,
0.1177/-0.0526, 0.0100/0.1089, -0.3081/0.3275, 0.6794/-0.1360, 0.0284/0.1835, 0.4033/-0.3811, -0.6592/0.1829, 0.0292/0.0417, 0.0752/-0.4210,
-0.4197/0.2743, 0.6959/-0.1412, -0.1376/-0.2561, -0.0945/0.2611, -0.1000/-0.2412, 0.2843/-0.3957, -0.6221/0.2155, -0.6234/0.2348, 0.5096/-0.3216,
0.6664/-0.1562, 0.6330/-0.2877, 0.3094/-0.3718, -0.2172/-0.1984, 0.6692/-0.2528, -0.4650/0.3045, 0.6302/-0.2273, 0.0303/-0.1818, 0.6435/-0.2694,
0.0835/-0.1584, 0.0719/0.1062, -0.0294/0.2352, -0.4181/0.3006, 0.6060/-0.2316, -0.1271/-0.2904, 0.0184/0.1191, -0.5012/0.2614, 0.7377/-0.1517,
-0.1172/0.2423, -0.1120/0.2557, 0.6242/-0.2618, 0.0888/-0.0380, -0.4396/0.3168, -0.1517/-0.2683, 0.4329/-0.3918, -0.1650/-0.2370, 0.0222/0.1978,
0.7206/-0.0839, 0.1193/0.0239, -0.2916/0.2804, -0.4070/0.3329, -0.0061/-0.2862, 0.0592/-0.0135, -0.1580/0.2596, 0.0739/-0.1352, 0.0570/0.1119,
0.6841/-0.0916, 0.4490/-0.4197, -0.3542/0.3205, -0.3186/0.2756, 0.1048/-0.0790, -0.0522/-0.3477, 0.1627/-0.3996, 0.0228/-0.1370, 0.0479/-0.3856,
0.6845/-0.1231, 0.7023/-0.0823, -0.0228/-0.3668, 0.1438/-0.3900, -0.0723/-0.2941, 0.0293/-0.4103, -0.2409/0.2804, 0.0453/-0.3740, 0.5261/-0.2543,
0.2605/-0.4217, 0.2443/-0.3814, 0.0878/-0.1697, 0.5428/-0.2980, 0.6347/-0.3031, -0.2251/0.2498, 0.7330/-0.0421, -0.0352/0.2022, 0.0604/0.1959,
0.2698/-0.3824, 0.1335/-0.0746, 0.6819/-0.1111, 0.4055/-0.4149, -0.0748/0.2215, -0.0280/-0.2615, -0.0148/0.0757, 0.7210/-0.1275, -0.0393/-0.3359,
0.6798/-0.2114, -0.1695/-0.2230, 0.7053/-0.1122, -0.0983/-0.3125, -0.1063/-0.2602, 0.0260/-0.2083, -0.4809/0.2403, -0.1245/-0.2565, -0.2445/0.2874,
-0.4699/0.2865, 0.4677/-0.3748, 0.0668/-0.3755, 0.1059/-0.3906, -0.1169/-0.3514, -0.3100/0.3218, -0.1705/0.2855, -0.1512/-0.1921, -0.5770/0.2266,
-0.3363/0.2802, 0.3245/-0.4328, -0.7202/0.1861, -0.0537/0.2622, -0.5156/0.2981, -0.0358/-0.3613, 0.6635/-0.1495, -0.1408/-0.2046, -0.3854/0.2996,
0.7103/-0.1408, -0.0138/0.2270, -0.0298/0.2192, -0.4264/0.2945, 0.0900/-0.1603, -0.6380/0.2174, 0.5766/-0.2567, -0.0986/0.2617, 0.5122/-0.3923,
0.6055/-0.2951, -0.6367/0.1776, -0.0740/0.2487, -0.1477/-0.2106, 0.3874/-0.3824, 0.4930/-0.3519, -0.2776/0.3260, 0.7143/-0.1828, 0.2457/-0.4429,
0.0460/0.1268, 0.0768/-0.2097, -0.2759/0.3119, 0.0885/0.0693, 0.0561/-0.3577, -0.1400/0.3070, 0.1329/-0.0618, 0.0522/0.1717, -0.6446/0.1984,
-0.2133/-0.0956, -0.2580/-0.1006, 0.5766/-0.2628, -0.0001/-0.3761, 0.4470/-0.3904, -0.4275/0.3047, -0.2149/0.3133, 0.0530/-0.1725, -0.5352/0.2785,
-0.0919/-0.2788, -0.0923/-0.2906, 0.1220/-0.1150, -0.1142/-0.2433, -0.1802/0.2779, 0.0831/-0.1455, -0.0470/0.2148, 0.6349/-0.2889, 0.6550/-0.2230,
0.7366/-0.0296, -0.0146/-0.2295, 0.3197/-0.3689, 0.6111/-0.2639, 0.0816/0.0739, 0.0630/-0.1868, 0.6739/-0.1496, 0.7244/-0.0745, 0.6771/-0.1516,
-0.2121/0.3226, -0.1801/-0.0929, 0.0379/-0.1962, 0.0162/-0.3533, -0.1181/0.2354, 0.0858/0.0392, 0.0996/-0.0594, 0.7294/-0.1300, 0.5194/-0.3205,
0.0681/0.0144, 0.0214/0.0930, 0.0778/-0.1699, -0.1591/-0.2601, 0.4367/-0.3425, 0.0988/0.0194, 0.6118/-0.2451, 0.0630/-0.1209, -0.1337/0.2962,
0.2150/-0.4237, 0.0806/-0.1985, 0.0367/-0.2349, -0.6896/0.1641, -0.1431/-0.2960, 0.0766/0.1089, -0.2011/-0.1326, -0.5882/0.2857, -0.5646/0.2559,
0.2781/-0.3990, 0.0569/-0.3770, -0.1855/-0.0919, 0.1117/-0.0551, -0.4403/0.3118, -0.2839/0.3045, -0.2427/-0.0637, -0.3747/0.3206, 0.0717/-0.0635,
0.1089/-0.4067, 0.6109/-0.1255, -0.0672/-0.3400, 0.0298/0.0484, 0.6637/-0.2671, 0.5217/-0.3488, -0.2405/0.2553, 0.1787/-0.4026, 0.7016/-0.0644,
-0.4352/0.2776, -0.0911/0.2604, -0.0953/0.2305, -0.0186/0.1817, -0.5394/0.2701
} {
  \fill[ALDGreen,opacity=.72] (\aldpx,\aldpy) circle[radius=.024];
}
\draw[ALDGreen,opacity=.45,line width=.45pt] plot[smooth] coordinates {(-0.0098,-0.2735) (0.0013,-0.2612) (0.0119,-0.2487) (0.0219,-0.2358) (0.0313,-0.2227) (0.0401,-0.2094) (0.0482,-0.1959) (0.0557,-0.1822) (0.0626,-0.1682) (0.0687,-0.1542) (0.0742,-0.1400) (0.0791,-0.1257) (0.0832,-0.1113) (0.0866,-0.0969) (0.0893,-0.0824) (0.0914,-0.0679) (0.0927,-0.0533) (0.0933,-0.0389) (0.0932,-0.0244) (0.0923,-0.0101) (0.0908,0.0042) (0.0886,0.0184) (0.0856,0.0324) (0.0820,0.0463) (0.0776,0.0600) (0.0726,0.0735) (0.0669,0.0868) (0.0605,0.0998) (0.0534,0.1126) (0.0457,0.1251) (0.0374,0.1373) (0.0284,0.1492) (0.0188,0.1607) (0.0086,0.1719) (-0.0021,0.1827) (-0.0135,0.1932) (-0.0254,0.2032) (-0.0378,0.2128) (-0.0507,0.2220) (-0.0641,0.2307) (-0.0780,0.2390) (-0.0923,0.2468) (-0.1071,0.2541) (-0.1222,0.2609) (-0.1378,0.2672) (-0.1537,0.2729) (-0.1699,0.2782) (-0.1865,0.2829) (-0.2033,0.2871) (-0.2204,0.2907) (-0.2377,0.2938) (-0.2552,0.2963) (-0.2729,0.2983) (-0.2907,0.2997) (-0.3087,0.3005) (-0.3268,0.3007) (-0.3449,0.3004) (-0.3630,0.2995) (-0.3812,0.2981) (-0.3994,0.2960) (-0.4175,0.2934) (-0.4355,0.2903) (-0.4534,0.2866) (-0.4713,0.2824) (-0.4889,0.2776) (-0.5064,0.2723) (-0.5236,0.2664) (-0.5406,0.2600) (-0.5574,0.2532) (-0.5739,0.2458) (-0.5900,0.2380) (-0.6058,0.2296) (-0.6213,0.2209) (-0.6363,0.2116) (-0.6510,0.2020) (-0.6652,0.1919) (-0.6790,0.1814) (-0.6923,0.1705) (-0.7051,0.1593) (-0.7173,0.1477)};
\draw[ALDGreen,opacity=.45,line width=.45pt] plot[smooth] coordinates {(0.7282,-0.0391) (0.7264,-0.0536) (0.7239,-0.0681) (0.7207,-0.0826) (0.7167,-0.0970) (0.7121,-0.1113) (0.7068,-0.1255) (0.7008,-0.1396) (0.6942,-0.1536) (0.6869,-0.1674) (0.6789,-0.1810) (0.6703,-0.1943) (0.6611,-0.2075) (0.6513,-0.2204) (0.6409,-0.2330) (0.6299,-0.2454) (0.6183,-0.2574) (0.6062,-0.2691) (0.5936,-0.2805) (0.5805,-0.2915) (0.5669,-0.3021) (0.5528,-0.3123) (0.5383,-0.3221) (0.5234,-0.3315) (0.5081,-0.3405) (0.4924,-0.3490) (0.4764,-0.3570) (0.4600,-0.3645) (0.4433,-0.3716) (0.4264,-0.3781) (0.4092,-0.3841) (0.3918,-0.3896) (0.3743,-0.3946) (0.3565,-0.3990) (0.3386,-0.4029) (0.3206,-0.4063) (0.3025,-0.4090) (0.2844,-0.4113) (0.2662,-0.4129) (0.2480,-0.4140) (0.2299,-0.4145) (0.2118,-0.4145) (0.1938,-0.4139) (0.1759,-0.4127) (0.1582,-0.4109) (0.1406,-0.4086) (0.1232,-0.4057) (0.1061,-0.4023) (0.0891,-0.3983) (0.0725,-0.3938) (0.0561,-0.3887) (0.0401,-0.3831) (0.0244,-0.3770) (0.0091,-0.3704) (-0.0058,-0.3632) (-0.0202,-0.3556) (-0.0343,-0.3475) (-0.0479,-0.3389) (-0.0610,-0.3299) (-0.0736,-0.3205) (-0.0856,-0.3106) (-0.0971,-0.3003) (-0.1081,-0.2896) (-0.1185,-0.2785) (-0.1283,-0.2671) (-0.1375,-0.2553) (-0.1461,-0.2433) (-0.1540,-0.2309) (-0.1613,-0.2182) (-0.1679,-0.2052) (-0.1738,-0.1920) (-0.1791,-0.1786) (-0.1837,-0.1650) (-0.1876,-0.1512) (-0.1908,-0.1372) (-0.1933,-0.1231) (-0.1951,-0.1088) (-0.1962,-0.0945) (-0.1965,-0.0801) (-0.1962,-0.0656)};
\end{scope}
\node[ald small] at (13.05,4.76) {$t=0$};
\node[text=ALDGreen] at (13.05,4.45) {$\sigma_0=0$};
\draw[ald arrow,draw=ALDMuted!65] (3.30,5.95) -- (4.05,5.95);
\draw[ald arrow,draw=ALDMuted!65] (7.05,5.95) -- (7.80,5.95);
\draw[ald arrow,draw=ALDMuted!65] (10.80,5.95) -- (11.55,5.95);

\node[anchor=east] at (1.36,3.91) {$\alpha(t)$};
\node[anchor=east] at (1.36,3.46) {$\sigma_t$};
\draw[ALDOrange,line width=1.15pt] plot coordinates {(1.8000,3.7300) (1.9125,3.7359) (2.0250,3.7419) (2.1375,3.7478) (2.2500,3.7538) (2.3625,3.7597) (2.4750,3.7656) (2.5875,3.7716) (2.7000,3.7775) (2.8125,3.7835) (2.9250,3.7894) (3.0375,3.7953) (3.1500,3.8013) (3.2625,3.8072) (3.3750,3.8132) (3.4875,3.8191) (3.6000,3.8250) (3.7125,3.8310) (3.8250,3.8369) (3.9375,3.8429) (4.0500,3.8488) (4.1625,3.8547) (4.2750,3.8607) (4.3875,3.8666) (4.5000,3.8726) (4.6125,3.8785) (4.7250,3.8844) (4.8375,3.8904) (4.9500,3.8963) (5.0625,3.9023) (5.1750,3.9082) (5.2875,3.9141) (5.4000,3.9201) (5.5125,3.9260) (5.5500,3.9280) (5.6250,3.9312) (5.7375,3.9361) (5.8500,3.9410) (5.9625,3.9458) (6.0750,3.9507) (6.1875,3.9555) (6.3000,3.9604) (6.4125,3.9653) (6.5250,3.9701) (6.6375,3.9750) (6.7500,3.9798) (6.8625,3.9847) (6.9750,3.9896) (7.0875,3.9944) (7.2000,3.9993) (7.3125,4.0041) (7.4250,4.0090) (7.5375,4.0139) (7.6500,4.0187) (7.7625,4.0236) (7.8750,4.0284) (7.9875,4.0333) (8.1000,4.0382) (8.2125,4.0430) (8.3250,4.0479) (8.4375,4.0527) (8.5500,4.0576) (8.6625,4.0625) (8.7750,4.0673) (8.8875,4.0722) (9.0000,4.0770) (9.1125,4.0819) (9.2250,4.0868) (9.3000,4.0900)};
\draw[ALDBlue,line width=1.15pt] plot coordinates {(9.3000,4.0900) (9.3375,4.0900) (9.4500,4.0900) (9.5625,4.0900) (9.6750,4.0900) (9.7875,4.0900) (9.9000,4.0900) (10.0125,4.0900) (10.1250,4.0900) (10.2375,4.0900) (10.3500,4.0900) (10.4625,4.0900) (10.5750,4.0900) (10.6875,4.0900) (10.8000,4.0900) (10.9125,4.0900) (11.0250,4.0900) (11.1375,4.0900) (11.2500,4.0900) (11.3625,4.0900) (11.4750,4.0900) (11.5875,4.0900) (11.7000,4.0900) (11.8125,4.0900) (11.9250,4.0900) (12.0375,4.0900) (12.1500,4.0900) (12.2625,4.0900) (12.3750,4.0900) (12.4875,4.0900) (12.6000,4.0900) (12.7125,4.0900) (12.8250,4.0900) (12.9375,4.0900) (13.0500,4.0900)};
\draw[ALDMuted,densely dashed,line width=.65pt] plot[smooth] coordinates {(1.8000,3.6000) (1.9125,3.5999) (2.0250,3.5998) (2.1375,3.5995) (2.2500,3.5991) (2.3625,3.5987) (2.4750,3.5981) (2.5875,3.5974) (2.7000,3.5966) (2.8125,3.5957) (2.9250,3.5946) (3.0375,3.5935) (3.1500,3.5923) (3.2625,3.5910) (3.3750,3.5895) (3.4875,3.5880) (3.6000,3.5864) (3.7125,3.5846) (3.8250,3.5828) (3.9375,3.5808) (4.0500,3.5788) (4.1625,3.5767) (4.2750,3.5744) (4.3875,3.5721) (4.5000,3.5697) (4.6125,3.5672) (4.7250,3.5645) (4.8375,3.5618) (4.9500,3.5590) (5.0625,3.5562) (5.1750,3.5532) (5.2875,3.5501) (5.4000,3.5470) (5.5125,3.5437) (5.5500,3.5426) (5.6250,3.5408) (5.7375,3.5380) (5.8500,3.5352) (5.9625,3.5323) (6.0750,3.5294) (6.1875,3.5264) (6.3000,3.5234) (6.4125,3.5203) (6.5250,3.5172) (6.6375,3.5140) (6.7500,3.5108) (6.8625,3.5075) (6.9750,3.5042) (7.0875,3.5009) (7.2000,3.4975) (7.3125,3.4940) (7.4250,3.4905) (7.5375,3.4870) (7.6500,3.4834) (7.7625,3.4798) (7.8750,3.4762) (7.9875,3.4725) (8.1000,3.4688) (8.2125,3.4650) (8.3250,3.4613) (8.4375,3.4574) (8.5500,3.4536) (8.6625,3.4497) (8.7750,3.4458) (8.8875,3.4419) (9.0000,3.4379) (9.1125,3.4339) (9.2250,3.4299) (9.3000,3.4273) (9.3375,3.4265) (9.4500,3.4243) (9.5625,3.4220) (9.6750,3.4197) (9.7875,3.4175) (9.9000,3.4152) (10.0125,3.4129) (10.1250,3.4106) (10.2375,3.4084) (10.3500,3.4061) (10.4625,3.4038) (10.5750,3.4015) (10.6875,3.3992) (10.8000,3.3968) (10.9125,3.3945) (11.0250,3.3922) (11.1375,3.3899) (11.2500,3.3876) (11.3625,3.3852) (11.4750,3.3829) (11.5875,3.3806) (11.7000,3.3782) (11.8125,3.3759) (11.9250,3.3735) (12.0375,3.3712) (12.1500,3.3688) (12.2625,3.3665) (12.3750,3.3641) (12.4875,3.3618) (12.6000,3.3594) (12.7125,3.3571) (12.8250,3.3547) (12.9375,3.3524) (13.0500,3.3500)};

\node[ald small,anchor=west] at (13.47,4.06) {tail shape};
\node[ald small,anchor=west] at (13.47,3.39) {noise scale};
\draw[ALDGrid,line width=.4pt] (0,3.07)--(15.25,3.07);

\node[ald heading] at (0,2.77)
  {(b) Heavy-tailed outside, Gaussian inside};
\node[ald small,anchor=east] at (15.25,2.77)
  {$\mathbf z_t=\sqrt{S_t}\,\boldsymbol\epsilon$};
\node[ald stage] at (1.85,2.35) {Heavy-tailed marginal};
\node[ald stage] at (6.94,2.35) {Condition on a scalar};
\node[ald stage] at (13.12,2.35) {Predict Gaussian noise};
\path[fill=ALDOrange,fill opacity=0.1] (0.3000,0.8800) -- plot[smooth] coordinates {(0.3000,0.8927) (0.3282,0.8933) (0.3564,0.8939) (0.3845,0.8945) (0.4127,0.8952) (0.4409,0.8960) (0.4691,0.8968) (0.4973,0.8976) (0.5255,0.8985) (0.5536,0.8995) (0.5818,0.9006) (0.6100,0.9018) (0.6382,0.9030) (0.6664,0.9044) (0.6945,0.9058) (0.7227,0.9074) (0.7509,0.9092) (0.7791,0.9111) (0.8073,0.9132) (0.8355,0.9155) (0.8636,0.9180) (0.8918,0.9208) (0.9200,0.9238) (0.9482,0.9272) (0.9764,0.9310) (1.0045,0.9352) (1.0327,0.9399) (1.0609,0.9452) (1.0891,0.9511) (1.1173,0.9577) (1.1455,0.9653) (1.1736,0.9738) (1.2018,0.9835) (1.2300,0.9946) (1.2582,1.0073) (1.2864,1.0218) (1.3145,1.0384) (1.3427,1.0574) (1.3709,1.0793) (1.3991,1.1044) (1.4273,1.1332) (1.4555,1.1659) (1.4836,1.2030) (1.5118,1.2448) (1.5400,1.2913) (1.5682,1.3423) (1.5964,1.3975) (1.6245,1.4559) (1.6527,1.5162) (1.6809,1.5764) (1.7091,1.6341) (1.7373,1.6868) (1.7655,1.7316) (1.7936,1.7658) (1.8218,1.7873) (1.8500,1.7947) (1.8782,1.7873) (1.9064,1.7658) (1.9345,1.7316) (1.9627,1.6868) (1.9909,1.6341) (2.0191,1.5764) (2.0473,1.5162) (2.0755,1.4559) (2.1036,1.3975) (2.1318,1.3423) (2.1600,1.2913) (2.1882,1.2448) (2.2164,1.2030) (2.2445,1.1659) (2.2727,1.1332) (2.3009,1.1044) (2.3291,1.0793) (2.3573,1.0574) (2.3855,1.0384) (2.4136,1.0218) (2.4418,1.0073) (2.4700,0.9946) (2.4982,0.9835) (2.5264,0.9738) (2.5545,0.9653) (2.5827,0.9577) (2.6109,0.9511) (2.6391,0.9452) (2.6673,0.9399) (2.6955,0.9352) (2.7236,0.9310) (2.7518,0.9272) (2.7800,0.9238) (2.8082,0.9208) (2.8364,0.9180) (2.8645,0.9155) (2.8927,0.9132) (2.9209,0.9111) (2.9491,0.9092) (2.9773,0.9074) (3.0055,0.9058) (3.0336,0.9044) (3.0618,0.9030) (3.0900,0.9018) (3.1182,0.9006) (3.1464,0.8995) (3.1745,0.8985) (3.2027,0.8976) (3.2309,0.8968) (3.2591,0.8960) (3.2873,0.8952) (3.3155,0.8945) (3.3436,0.8939) (3.3718,0.8933) (3.4000,0.8927)} -- (3.4000,0.8800) -- cycle;
\draw[ALDOrange,line width=.85pt] plot[smooth] coordinates {(0.3000,0.8927) (0.3282,0.8933) (0.3564,0.8939) (0.3845,0.8945) (0.4127,0.8952) (0.4409,0.8960) (0.4691,0.8968) (0.4973,0.8976) (0.5255,0.8985) (0.5536,0.8995) (0.5818,0.9006) (0.6100,0.9018) (0.6382,0.9030) (0.6664,0.9044) (0.6945,0.9058) (0.7227,0.9074) (0.7509,0.9092) (0.7791,0.9111) (0.8073,0.9132) (0.8355,0.9155) (0.8636,0.9180) (0.8918,0.9208) (0.9200,0.9238) (0.9482,0.9272) (0.9764,0.9310) (1.0045,0.9352) (1.0327,0.9399) (1.0609,0.9452) (1.0891,0.9511) (1.1173,0.9577) (1.1455,0.9653) (1.1736,0.9738) (1.2018,0.9835) (1.2300,0.9946) (1.2582,1.0073) (1.2864,1.0218) (1.3145,1.0384) (1.3427,1.0574) (1.3709,1.0793) (1.3991,1.1044) (1.4273,1.1332) (1.4555,1.1659) (1.4836,1.2030) (1.5118,1.2448) (1.5400,1.2913) (1.5682,1.3423) (1.5964,1.3975) (1.6245,1.4559) (1.6527,1.5162) (1.6809,1.5764) (1.7091,1.6341) (1.7373,1.6868) (1.7655,1.7316) (1.7936,1.7658) (1.8218,1.7873) (1.8500,1.7947) (1.8782,1.7873) (1.9064,1.7658) (1.9345,1.7316) (1.9627,1.6868) (1.9909,1.6341) (2.0191,1.5764) (2.0473,1.5162) (2.0755,1.4559) (2.1036,1.3975) (2.1318,1.3423) (2.1600,1.2913) (2.1882,1.2448) (2.2164,1.2030) (2.2445,1.1659) (2.2727,1.1332) (2.3009,1.1044) (2.3291,1.0793) (2.3573,1.0574) (2.3855,1.0384) (2.4136,1.0218) (2.4418,1.0073) (2.4700,0.9946) (2.4982,0.9835) (2.5264,0.9738) (2.5545,0.9653) (2.5827,0.9577) (2.6109,0.9511) (2.6391,0.9452) (2.6673,0.9399) (2.6955,0.9352) (2.7236,0.9310) (2.7518,0.9272) (2.7800,0.9238) (2.8082,0.9208) (2.8364,0.9180) (2.8645,0.9155) (2.8927,0.9132) (2.9209,0.9111) (2.9491,0.9092) (2.9773,0.9074) (3.0055,0.9058) (3.0336,0.9044) (3.0618,0.9030) (3.0900,0.9018) (3.1182,0.9006) (3.1464,0.8995) (3.1745,0.8985) (3.2027,0.8976) (3.2309,0.8968) (3.2591,0.8960) (3.2873,0.8952) (3.3155,0.8945) (3.3436,0.8939) (3.3718,0.8933) (3.4000,0.8927)};

\draw[ALDGrid,line width=.35pt] (.22,.88)--(3.48,.88);
\draw[ALDGrid,densely dashed,line width=.35pt] (1.85,.88)--(1.85,2.03);
\node[ald small] at (1.85,.66) {$m_t\mathbf x_0$};
\node[text=ALDOrange,anchor=west] at (.32,1.43) {long tails};
\draw[ald arrow,draw=ALDOrange,line width=.45pt]
  (.96,1.32)--(.70,1.03);
\node at (1.85,.30) {$q(\mathbf x_t\mid\mathbf x_0)$};
\draw[ald arrow] (3.67,1.29)--(4.17,1.29);
\path[fill=ALDBlue,fill opacity=0.065] (4.7300,1.0300) -- plot[smooth] coordinates {(4.7300,1.0431) (4.7666,1.0447) (4.8032,1.0464) (4.8398,1.0482) (4.8764,1.0503) (4.9130,1.0525) (4.9496,1.0548) (4.9862,1.0574) (5.0228,1.0601) (5.0594,1.0630) (5.0960,1.0662) (5.1326,1.0695) (5.1691,1.0730) (5.2057,1.0768) (5.2423,1.0807) (5.2789,1.0849) (5.3155,1.0892) (5.3521,1.0937) (5.3887,1.0984) (5.4253,1.1033) (5.4619,1.1083) (5.4985,1.1135) (5.5351,1.1188) (5.5717,1.1242) (5.6083,1.1297) (5.6449,1.1353) (5.6815,1.1409) (5.7181,1.1465) (5.7547,1.1522) (5.7913,1.1577) (5.8279,1.1632) (5.8645,1.1687) (5.9011,1.1739) (5.9377,1.1791) (5.9743,1.1840) (6.0109,1.1887) (6.0474,1.1932) (6.0840,1.1974) (6.1206,1.2013) (6.1572,1.2048) (6.1938,1.2080) (6.2304,1.2108) (6.2670,1.2133) (6.3036,1.2153) (6.3402,1.2168) (6.3768,1.2180) (6.4134,1.2186) (6.4500,1.2189) (6.4866,1.2186) (6.5232,1.2180) (6.5598,1.2168) (6.5964,1.2153) (6.6330,1.2133) (6.6696,1.2108) (6.7062,1.2080) (6.7428,1.2048) (6.7794,1.2013) (6.8160,1.1974) (6.8526,1.1932) (6.8891,1.1887) (6.9257,1.1840) (6.9623,1.1791) (6.9989,1.1739) (7.0355,1.1687) (7.0721,1.1632) (7.1087,1.1577) (7.1453,1.1522) (7.1819,1.1465) (7.2185,1.1409) (7.2551,1.1353) (7.2917,1.1297) (7.3283,1.1242) (7.3649,1.1188) (7.4015,1.1135) (7.4381,1.1083) (7.4747,1.1033) (7.5113,1.0984) (7.5479,1.0937) (7.5845,1.0892) (7.6211,1.0849) (7.6577,1.0807) (7.6943,1.0768) (7.7309,1.0730) (7.7674,1.0695) (7.8040,1.0662) (7.8406,1.0630) (7.8772,1.0601) (7.9138,1.0574) (7.9504,1.0548) (7.9870,1.0525) (8.0236,1.0503) (8.0602,1.0482) (8.0968,1.0464) (8.1334,1.0447) (8.1700,1.0431)} -- (8.1700,1.0300) -- cycle;
\draw[ALDBlue,line width=.85pt] plot[smooth] coordinates {(4.7300,1.0431) (4.7666,1.0447) (4.8032,1.0464) (4.8398,1.0482) (4.8764,1.0503) (4.9130,1.0525) (4.9496,1.0548) (4.9862,1.0574) (5.0228,1.0601) (5.0594,1.0630) (5.0960,1.0662) (5.1326,1.0695) (5.1691,1.0730) (5.2057,1.0768) (5.2423,1.0807) (5.2789,1.0849) (5.3155,1.0892) (5.3521,1.0937) (5.3887,1.0984) (5.4253,1.1033) (5.4619,1.1083) (5.4985,1.1135) (5.5351,1.1188) (5.5717,1.1242) (5.6083,1.1297) (5.6449,1.1353) (5.6815,1.1409) (5.7181,1.1465) (5.7547,1.1522) (5.7913,1.1577) (5.8279,1.1632) (5.8645,1.1687) (5.9011,1.1739) (5.9377,1.1791) (5.9743,1.1840) (6.0109,1.1887) (6.0474,1.1932) (6.0840,1.1974) (6.1206,1.2013) (6.1572,1.2048) (6.1938,1.2080) (6.2304,1.2108) (6.2670,1.2133) (6.3036,1.2153) (6.3402,1.2168) (6.3768,1.2180) (6.4134,1.2186) (6.4500,1.2189) (6.4866,1.2186) (6.5232,1.2180) (6.5598,1.2168) (6.5964,1.2153) (6.6330,1.2133) (6.6696,1.2108) (6.7062,1.2080) (6.7428,1.2048) (6.7794,1.2013) (6.8160,1.1974) (6.8526,1.1932) (6.8891,1.1887) (6.9257,1.1840) (6.9623,1.1791) (6.9989,1.1739) (7.0355,1.1687) (7.0721,1.1632) (7.1087,1.1577) (7.1453,1.1522) (7.1819,1.1465) (7.2185,1.1409) (7.2551,1.1353) (7.2917,1.1297) (7.3283,1.1242) (7.3649,1.1188) (7.4015,1.1135) (7.4381,1.1083) (7.4747,1.1033) (7.5113,1.0984) (7.5479,1.0937) (7.5845,1.0892) (7.6211,1.0849) (7.6577,1.0807) (7.6943,1.0768) (7.7309,1.0730) (7.7674,1.0695) (7.8040,1.0662) (7.8406,1.0630) (7.8772,1.0601) (7.9138,1.0574) (7.9504,1.0548) (7.9870,1.0525) (8.0236,1.0503) (8.0602,1.0482) (8.0968,1.0464) (8.1334,1.0447) (8.1700,1.0431)};
\path[fill=ALDGreen,fill opacity=0.065] (4.7300,1.1800) -- plot[smooth] coordinates {(4.7300,1.1801) (4.7666,1.1802) (4.8032,1.1802) (4.8398,1.1803) (4.8764,1.1804) (4.9130,1.1805) (4.9496,1.1807) (4.9862,1.1810) (5.0228,1.1813) (5.0594,1.1818) (5.0960,1.1823) (5.1326,1.1830) (5.1691,1.1839) (5.2057,1.1850) (5.2423,1.1863) (5.2789,1.1880) (5.3155,1.1901) (5.3521,1.1926) (5.3887,1.1956) (5.4253,1.1991) (5.4619,1.2033) (5.4985,1.2083) (5.5351,1.2140) (5.5717,1.2206) (5.6083,1.2282) (5.6449,1.2367) (5.6815,1.2462) (5.7181,1.2568) (5.7547,1.2685) (5.7913,1.2812) (5.8279,1.2949) (5.8645,1.3094) (5.9011,1.3248) (5.9377,1.3409) (5.9743,1.3574) (6.0109,1.3742) (6.0474,1.3911) (6.0840,1.4077) (6.1206,1.4240) (6.1572,1.4395) (6.1938,1.4539) (6.2304,1.4671) (6.2670,1.4788) (6.3036,1.4887) (6.3402,1.4966) (6.3768,1.5024) (6.4134,1.5060) (6.4500,1.5071) (6.4866,1.5060) (6.5232,1.5024) (6.5598,1.4966) (6.5964,1.4887) (6.6330,1.4788) (6.6696,1.4671) (6.7062,1.4539) (6.7428,1.4395) (6.7794,1.4240) (6.8160,1.4077) (6.8526,1.3911) (6.8891,1.3742) (6.9257,1.3574) (6.9623,1.3409) (6.9989,1.3248) (7.0355,1.3094) (7.0721,1.2949) (7.1087,1.2812) (7.1453,1.2685) (7.1819,1.2568) (7.2185,1.2462) (7.2551,1.2367) (7.2917,1.2282) (7.3283,1.2206) (7.3649,1.2140) (7.4015,1.2083) (7.4381,1.2033) (7.4747,1.1991) (7.5113,1.1956) (7.5479,1.1926) (7.5845,1.1901) (7.6211,1.1880) (7.6577,1.1863) (7.6943,1.1850) (7.7309,1.1839) (7.7674,1.1830) (7.8040,1.1823) (7.8406,1.1818) (7.8772,1.1813) (7.9138,1.1810) (7.9504,1.1807) (7.9870,1.1805) (8.0236,1.1804) (8.0602,1.1803) (8.0968,1.1802) (8.1334,1.1802) (8.1700,1.1801)} -- (8.1700,1.1800) -- cycle;
\draw[ALDGreen,line width=.85pt] plot[smooth] coordinates {(4.7300,1.1801) (4.7666,1.1802) (4.8032,1.1802) (4.8398,1.1803) (4.8764,1.1804) (4.9130,1.1805) (4.9496,1.1807) (4.9862,1.1810) (5.0228,1.1813) (5.0594,1.1818) (5.0960,1.1823) (5.1326,1.1830) (5.1691,1.1839) (5.2057,1.1850) (5.2423,1.1863) (5.2789,1.1880) (5.3155,1.1901) (5.3521,1.1926) (5.3887,1.1956) (5.4253,1.1991) (5.4619,1.2033) (5.4985,1.2083) (5.5351,1.2140) (5.5717,1.2206) (5.6083,1.2282) (5.6449,1.2367) (5.6815,1.2462) (5.7181,1.2568) (5.7547,1.2685) (5.7913,1.2812) (5.8279,1.2949) (5.8645,1.3094) (5.9011,1.3248) (5.9377,1.3409) (5.9743,1.3574) (6.0109,1.3742) (6.0474,1.3911) (6.0840,1.4077) (6.1206,1.4240) (6.1572,1.4395) (6.1938,1.4539) (6.2304,1.4671) (6.2670,1.4788) (6.3036,1.4887) (6.3402,1.4966) (6.3768,1.5024) (6.4134,1.5060) (6.4500,1.5071) (6.4866,1.5060) (6.5232,1.5024) (6.5598,1.4966) (6.5964,1.4887) (6.6330,1.4788) (6.6696,1.4671) (6.7062,1.4539) (6.7428,1.4395) (6.7794,1.4240) (6.8160,1.4077) (6.8526,1.3911) (6.8891,1.3742) (6.9257,1.3574) (6.9623,1.3409) (6.9989,1.3248) (7.0355,1.3094) (7.0721,1.2949) (7.1087,1.2812) (7.1453,1.2685) (7.1819,1.2568) (7.2185,1.2462) (7.2551,1.2367) (7.2917,1.2282) (7.3283,1.2206) (7.3649,1.2140) (7.4015,1.2083) (7.4381,1.2033) (7.4747,1.1991) (7.5113,1.1956) (7.5479,1.1926) (7.5845,1.1901) (7.6211,1.1880) (7.6577,1.1863) (7.6943,1.1850) (7.7309,1.1839) (7.7674,1.1830) (7.8040,1.1823) (7.8406,1.1818) (7.8772,1.1813) (7.9138,1.1810) (7.9504,1.1807) (7.9870,1.1805) (8.0236,1.1804) (8.0602,1.1803) (8.0968,1.1802) (8.1334,1.1802) (8.1700,1.1801)};
\path[fill=ALDOrange,fill opacity=0.065] (4.7300,1.3300) -- plot[smooth] coordinates {(4.7300,1.3300) (4.7666,1.3300) (4.8032,1.3300) (4.8398,1.3300) (4.8764,1.3300) (4.9130,1.3300) (4.9496,1.3300) (4.9862,1.3300) (5.0228,1.3300) (5.0594,1.3300) (5.0960,1.3300) (5.1326,1.3300) (5.1691,1.3300) (5.2057,1.3300) (5.2423,1.3300) (5.2789,1.3300) (5.3155,1.3300) (5.3521,1.3300) (5.3887,1.3300) (5.4253,1.3300) (5.4619,1.3301) (5.4985,1.3302) (5.5351,1.3303) (5.5717,1.3306) (5.6083,1.3310) (5.6449,1.3317) (5.6815,1.3329) (5.7181,1.3348) (5.7547,1.3376) (5.7913,1.3420) (5.8279,1.3482) (5.8645,1.3572) (5.9011,1.3695) (5.9377,1.3861) (5.9743,1.4077) (6.0109,1.4350) (6.0474,1.4686) (6.0840,1.5086) (6.1206,1.5546) (6.1572,1.6058) (6.1938,1.6606) (6.2304,1.7167) (6.2670,1.7717) (6.3036,1.8224) (6.3402,1.8658) (6.3768,1.8991) (6.4134,1.9201) (6.4500,1.9273) (6.4866,1.9201) (6.5232,1.8991) (6.5598,1.8658) (6.5964,1.8224) (6.6330,1.7717) (6.6696,1.7167) (6.7062,1.6606) (6.7428,1.6058) (6.7794,1.5546) (6.8160,1.5086) (6.8526,1.4686) (6.8891,1.4350) (6.9257,1.4077) (6.9623,1.3861) (6.9989,1.3695) (7.0355,1.3572) (7.0721,1.3482) (7.1087,1.3420) (7.1453,1.3376) (7.1819,1.3348) (7.2185,1.3329) (7.2551,1.3317) (7.2917,1.3310) (7.3283,1.3306) (7.3649,1.3303) (7.4015,1.3302) (7.4381,1.3301) (7.4747,1.3300) (7.5113,1.3300) (7.5479,1.3300) (7.5845,1.3300) (7.6211,1.3300) (7.6577,1.3300) (7.6943,1.3300) (7.7309,1.3300) (7.7674,1.3300) (7.8040,1.3300) (7.8406,1.3300) (7.8772,1.3300) (7.9138,1.3300) (7.9504,1.3300) (7.9870,1.3300) (8.0236,1.3300) (8.0602,1.3300) (8.0968,1.3300) (8.1334,1.3300) (8.1700,1.3300)} -- (8.1700,1.3300) -- cycle;
\draw[ALDOrange,line width=.85pt] plot[smooth] coordinates {(4.7300,1.3300) (4.7666,1.3300) (4.8032,1.3300) (4.8398,1.3300) (4.8764,1.3300) (4.9130,1.3300) (4.9496,1.3300) (4.9862,1.3300) (5.0228,1.3300) (5.0594,1.3300) (5.0960,1.3300) (5.1326,1.3300) (5.1691,1.3300) (5.2057,1.3300) (5.2423,1.3300) (5.2789,1.3300) (5.3155,1.3300) (5.3521,1.3300) (5.3887,1.3300) (5.4253,1.3300) (5.4619,1.3301) (5.4985,1.3302) (5.5351,1.3303) (5.5717,1.3306) (5.6083,1.3310) (5.6449,1.3317) (5.6815,1.3329) (5.7181,1.3348) (5.7547,1.3376) (5.7913,1.3420) (5.8279,1.3482) (5.8645,1.3572) (5.9011,1.3695) (5.9377,1.3861) (5.9743,1.4077) (6.0109,1.4350) (6.0474,1.4686) (6.0840,1.5086) (6.1206,1.5546) (6.1572,1.6058) (6.1938,1.6606) (6.2304,1.7167) (6.2670,1.7717) (6.3036,1.8224) (6.3402,1.8658) (6.3768,1.8991) (6.4134,1.9201) (6.4500,1.9273) (6.4866,1.9201) (6.5232,1.8991) (6.5598,1.8658) (6.5964,1.8224) (6.6330,1.7717) (6.6696,1.7167) (6.7062,1.6606) (6.7428,1.6058) (6.7794,1.5546) (6.8160,1.5086) (6.8526,1.4686) (6.8891,1.4350) (6.9257,1.4077) (6.9623,1.3861) (6.9989,1.3695) (7.0355,1.3572) (7.0721,1.3482) (7.1087,1.3420) (7.1453,1.3376) (7.1819,1.3348) (7.2185,1.3329) (7.2551,1.3317) (7.2917,1.3310) (7.3283,1.3306) (7.3649,1.3303) (7.4015,1.3302) (7.4381,1.3301) (7.4747,1.3300) (7.5113,1.3300) (7.5479,1.3300) (7.5845,1.3300) (7.6211,1.3300) (7.6577,1.3300) (7.6943,1.3300) (7.7309,1.3300) (7.7674,1.3300) (7.8040,1.3300) (7.8406,1.3300) (7.8772,1.3300) (7.9138,1.3300) (7.9504,1.3300) (7.9870,1.3300) (8.0236,1.3300) (8.0602,1.3300) (8.0968,1.3300) (8.1334,1.3300) (8.1700,1.3300)};

\draw[ALDMuted!65,densely dashed,line width=.4pt]
  (6.45,.91)--(6.45,2.08);
\node[ald small] at (6.45,.72) {shared mean};
\node[text=ALDOrange,anchor=west] at (8.36,1.79) {$S_t=0.3$};
\node[text=ALDGreen,anchor=west] at (8.36,1.45) {$S_t=1$};
\node[text=ALDBlue,anchor=west] at (8.36,1.11) {$S_t=3$};
\node at (6.94,.30)
  {$q(\mathbf x_t\mid\mathbf x_0,S_t=s)
     =\mathcal N(m_t\mathbf x_0,\sigma_t^2s\mathbf I)$};
\draw[ald arrow] (9.75,1.29)--(10.30,1.29);

\node at (12.20,1.95) {$\mathbf x_t,\ t,\ \log S_t$};
\draw[ald arrow,draw=ALDBlue] (12.20,1.73)--(12.20,1.49);
\node[draw=ALDBlue,fill=ALDBlue!6,rounded corners=2pt,
      minimum width=1.72cm,minimum height=.66cm,
      line width=.75pt,text=ALDBlue,
      font=\sffamily\fontsize{15}{17}\selectfont]
  (aldpredictor) at (12.20,1.13) {$\epsilon_\theta$};
\draw[ald arrow,draw=ALDBlue] (aldpredictor.east)--(13.69,1.13);
\node[text=ALDBlue,font=\sffamily\fontsize{12}{14}\selectfont]
  at (14.10,1.13) {$\widehat{\boldsymbol\epsilon}$};
\node[ald small] at (12.83,.43)
  {$\alpha(t)=2\ \Longrightarrow\ S_t\equiv1$};

\end{tikzpicture}%
}
\endgroup

\caption{ALD combines tail annealing with a Gaussian scale-mixture
representation. (a) Illustrative perturbation marginals along generation:
the stability index $\alpha(t)$ reaches $2$ at $t=\tau$, while $\sigma_t$
separately controls noise magnitude. (b) Random $S_t$ yields heavy-tailed
marginal noise, while conditioning on $\mathbf x_0$ and $S_t$ gives a
Gaussian kernel and enables conditional $\epsilon$-prediction.
At $\alpha(t)=2$, $S_t\equiv1$.
The slices show one-time marginals, not a cross-time sampling coupling;
the kernel curves depict the ideal unclamped construction.}
\label{fig:ald_ac}
\end{figure}

\paragraph{Schedule for time-varying $\alpha(t)$.}
Intuitively, we use $\alpha(t)$ to establish a two-phase denoising dynamic. 
In the early exploration phase ($t>\tau$), heavy-tailed perturbations
are intended to facilitate exploration across low-density inter-mode
regions.
In the subsequent refinement phase ($t\leq\tau$), the perturbation
becomes Gaussian ($S_t\equiv1$), reducing large noise impulses near
the data manifold. To realize this mechanism, we adopt a piecewise linear schedule (Figure~\ref{fig:ald_ac}):
\begin{equation}
  \alpha(t) =
  \begin{cases}
    2 - \dfrac{(2 - \alpha_{\min})(t - \tau)}{1 - \tau}, & t \geq \tau, \\[6pt]
    2, & t < \tau,
  \end{cases}
  \label{eq:alpha_schedule}
\end{equation}
where $\tau \in [0, 1)$ denotes the phase-transition threshold and $\alpha_{\min} \in (1, 2)$ bounds the tail heaviness. As demonstrated in Section~\ref{sec:experiments}, this schedule yields clear empirical benefits:
\begin{itemize}
  \item On light-tailed and manifold heavy-tailed benchmarks, ALD matches the performance of DDPM, while the DLPM may exhibit a heavy-tail hallucination (Section~\ref{exp:synthetic}).
  \item On globally heavy-tailed distributions, ALD strictly outperforms DDPM while incurring substantially lower training costs (Sections~\ref{exp:synthetic}, \ref{exp:holtsmark_robustness}).
  \item Across all evaluated settings, ALD consistently achieves higher sample fidelity than DLPM (Section~\ref{exp:synthetic}).
\end{itemize}

\paragraph{Conditional Gaussianity.}
Conditioned on $S_t$, the forward kernel is Gaussian:
$\bx_t \mid \bx_0, S_t \sim \mathcal{N}(m_t \bx_0,\, \sigma_t^2 S_t \mathbf{I})$.
This is the key computational property:
all heavy-tailed behavior is encoded in the scalar $S_t$,
while the conditional structure remains analytically tractable and identical
to standard score matching.

\subsection{Sampling Process and Training Objective}
\label{sec:method:flow}

The forward construction specifies the perturbation distribution
at each time, but does not determine transitions between times.
We therefore introduce two sampling rules below. We first define the following two transitions using their
ideal conditional means. The network approximation is introduced only in the
training paragraph below.

\paragraph{Direct marginal re-noising.}
For the marginal state $(\bx_t, S_t)$, define the ALD-renoise denoiser
\begin{equation}
 \hat\bx_{0}^{\mathrm R}
 :=\mathbb E[\bx_0\mid\bx_t, S_t]
 =\frac{\bx_t-\sigma_t\sqrt{ S_t}\,
   \mathbb E[\bn\mid\bx_t, S_t]}{m_t},
 \qquad 0<t<1.
 \label{eq:tweedie}
\end{equation}
The sampler independently draws $ S_{t-\Delta t}$ from the prescribed
next-time scale law and
$\bn'\sim\mathcal N(\mathbf0,\mathbf I)$, then updates
\begin{equation}
 \bx_{t-\Delta t}
 =m_{t-\Delta t}\hat\bx_0^{\mathrm R}
  +\sigma_{t-\Delta t}\sqrt{ S_{t-\Delta t}}\,\bn'.
 \label{eq:ald_renoise_step}
\end{equation}
We call Eq.~\eqref{eq:ald_renoise_step} \emph{ALD-renoise}; its complete
implementation is given in Algorithm~\ref{alg:ald}. The update uses the current
state only through $\hat\bx_0^{\mathrm R}$ and injects fresh noise at the full
next-time marginal scale. It therefore does not preserve the local dependence
between adjacent states. Repeating full marginal re-noising can over-inject
randomness, producing oscillatory or unstable finite-step trajectories and
amplifying reconstruction errors.

At $t=1$, $m_1=0$, so Eq.~\eqref{eq:tweedie} is not evaluated. Independence
gives $\mathbb E[\bx_0\mid\bx_1, S_1]=\mathbb E[\bx_0]$, for which we use
the empirical estimate
$\bar\bx_{\mathrm{data}}:=M^{-1}\sum_{j=1}^{M}\bx_0^{(j)}$.

\paragraph{Variance-controlled adjacent transition.}
In our high-dimensional image experiments , we observed unstable generation with ALD-renoise.
Its update injects fresh noise at the full next-time marginal scale
after each denoising step.
This motivates controlling the variance injected at each reverse step
to improve sampling stability.

Following the Gaussian forward transition in DDPM
\citep{ho2020ddpm}, we construct an adjacent transition using
our coefficients $m_t$ and $\sigma_t$.
Scaling $\bx_{t-\Delta t}$ by $m_t/m_{t-\Delta t}$ leaves
a noise variance of
$(m_t^2/m_{t-\Delta t}^2)\sigma_{t-\Delta t}^2$.
To obtain
$\operatorname{Cov}(\bx_t\mid\bx_0)=\sigma_t^2\mathbf I$,
we add independent Gaussian noise with variance
$\sigma_t^2-(m_t^2/m_{t-\Delta t}^2)\sigma_{t-\Delta t}^2$.
We multiply this variance by the current scale $ S_t$, giving
\begin{equation}
 \bx_t
 =\frac{m_t}{m_{t-\Delta t}}\bx_{t-\Delta t}
   +\sqrt{\left(\sigma_t^2-
      \frac{m_t^2}{m_{t-\Delta t}^2}\sigma_{t-\Delta t}^2\right)
        S_t}\,\bn_t,
 \qquad
 \bn_t\sim\mathcal N(\mathbf0,\mathbf I).
 \label{eq:ald_posterior_forward}
\end{equation}
At each step, independently draw
$S_t\sim\mathcal A_{\alpha(t)/2}$ .
For $\alpha(t)=2$, set $ S_t=1$.

Let $V_t$ denote the accumulated conditional noise variance.
Independence of the added noise gives
\begin{equation}
 V_t
 =\frac{m_t^2}{m_{t-\Delta t}^2}V_{t-\Delta t}
   +\left(\sigma_t^2-
      \frac{m_t^2}{m_{t-\Delta t}^2}\sigma_{t-\Delta t}^2\right)
     S_t,
 \qquad V_0=0.
 \label{eq:ald_posterior_variance}
\end{equation}
Thus, $S_t$ controls the newly added noise, whereas $V_t$
accounts for the noise accumulated across preceding steps.
Conditioned on $\bx_0$ and $V_t$, the current state satisfies
$\bx_t\mid\bx_0,V_t\sim\mathcal N(m_t\bx_0,V_t\mathbf I)$.Similarly to
\citep{ho2020ddpm}, we construct the adjacent posterior for the
forward coupling in Eq.~\eqref{eq:ald_posterior_forward}.
Gaussian conditioning yields
\begin{align}
 \mathbb E[\bx_{t-\Delta t}\mid\bx_t,\bx_0,V_{t-\Delta t},V_t]
 &=\frac{m_tV_{t-\Delta t}}{m_{t-\Delta t}V_t}\bx_t
   +m_{t-\Delta t}\left(
      1-\frac{m_t^2V_{t-\Delta t}}
              {m_{t-\Delta t}^2V_t}\right)\bx_0,
 \label{eq:ald_posterior_mean}\\
 \operatorname{Cov}(\bx_{t-\Delta t}\mid\bx_t,\bx_0,
                    V_{t-\Delta t},V_t)
 &=V_{t-\Delta t}\left(
      1-\frac{m_t^2V_{t-\Delta t}}
              {m_{t-\Delta t}^2V_t}\right)\mathbf I.
 \label{eq:ald_posterior_covariance}
\end{align}

For $0<t<1$, the corresponding denoiser is
\begin{equation}
 \hat\bx_0^{\mathrm P}
 :=\mathbb E[\bx_0\mid\bx_t,V_t]
 =\frac{\bx_t-\sqrt{V_t}\,
   \mathbb E[\bn\mid\bx_t,V_t]}{m_t}.
 \label{eq:ald_posterior_denoiser}
\end{equation}
Replacing $\bx_0$ in Eq.~\eqref{eq:ald_posterior_mean} by the denoiser
$\hat\bx_0^{\mathrm P}$ and using the covariance in
Eq.~\eqref{eq:ald_posterior_covariance} defines \emph{ALD-posterior}, a Gaussian approximation to the
reverse conditional.
This update injects only the adjacent conditional variance,
which is no greater than $V_{t-\Delta t}$, rather than fresh noise
at the full next-time marginal scale.
Algorithm~\ref{alg:ald_posterior} gives the sampling procedure.

However, this construction changes the forward marginal distributions.
The accumulated variance $V_t$ depends on the scales sampled
across steps, whereas the original perturbation uses
$\sigma_t^2S_t$ at each individual time.
Consequently, this construction generally changes the forward
marginal distribution of $\bx_t$ relative to
Eq.~\eqref{eq:marginal_forward}.
Nevertheless, the alternative forward process still starts from
$p_{\mathrm{data}}$, since $m_0=1$ and $V_0=0$.
Our objective remains generation from this data distribution,
using the modified perturbation family to obtain a more stable
sampling procedure.

\begin{remark}[Preserving the Original Marginals]
\label{rem:marginal_variance_constraint}
To preserve the original conditional noise variance $\sigma_t^2S_t$
under an adjacent linear Gaussian transition, the variance of the
added noise must be
\[
 \sigma_t^2S_t
 -\frac{m_t^2}{m_{t-\Delta t}^2}
  \sigma_{t-\Delta t}^2S_{t-\Delta t}.
\]
A valid Gaussian transition therefore requires
\begin{equation*}
    \sigma_t^2S_t
 \geq
 \frac{m_t^2}{m_{t-\Delta t}^2}
 \sigma_{t-\Delta t}^2S_{t-\Delta t}.
\end{equation*}
This condition need not hold for independently sampled scales.
If it fails, both the added forward noise variance and the
corresponding reverse conditional variance become negative.
The accumulated-variance construction in
Eq.~\eqref{eq:ald_posterior_variance} avoids this issue, but generally
changes the original forward marginal distributions.
\end{remark}

\begin{remark}[Implementation]
\label{rem:posterior_implementation}
Although ALD-posterior changes the forward marginal distributions,
the perturbation remains Gaussian conditioned on its noise scale.
ALD-posterior can therefore use the same form of
$\epsilon$-prediction objective as ALD-renoise, with
$V_t/\sigma_t^2$ as the scale-conditioning input.
The training objective is described below.
\end{remark}

\paragraph{Training objective.}
To implement the denoiser in
Eqs.~\eqref{eq:tweedie} and~\eqref{eq:ald_posterior_denoiser}, we use the same
noise predictor at different scale inputs. For ALD-renoise,
$\mathbb E[\bn\mid\bx_t,\tilde S_t]$ is approximated by
$\eps_{\btheta}(\bx_t,t,\log(\tilde S_t+10^{-6}))$; for ALD-posterior,
$\mathbb E[\bn\mid\bx_t,V_t]$ is approximated by
$\eps_{\btheta}(\bx_t,t,\log(V_t/\sigma_t^2+10^{-6}))$. Consequently, we present the training formulation
for ALD-renoise only.  For numerical stability, we use the truncated scale
$\tilde S_t:=\min(S_t,S_{\max})$ in both training and sampling.
We train
$\eps_{\btheta}(\bx_t,t,\log(\tilde S_t+10^{-6}))$
to estimate $\mathbb E[\bn\mid\bx_t,\tilde S_t]$ by Gaussian noise prediction:
\begin{equation}
  \mathcal{L}_{\mathrm{DSM}}(\btheta)
  =
  \E_{t,\,\bx_0,\,S_t,\,\bn}\!\Bigl[
    \frac{1}{\sigma_t^2\tilde{S}_t}
    \Bigl\|\eps_{\btheta}\bigl(m_t\bx_0+\sigma_t\sqrt{\tilde{S}_t}\,\bn,\;
    t,\;\log(\tilde{S}_t+10^{-6})\bigr) - \bn\Bigr\|_2^2 \Bigr].
  \label{eq:weighted_dsm_loss}
\end{equation}
To avoid unstable overweighting of small-noise states, an unweighted $\epsilon$-prediction objective is used:
\begin{equation}
  \mathcal{L}(\btheta)
  = \E_{t,\,\bx_0,\,S_t,\,\bn}\!\Bigl[
    \Bigl\|\eps_{\btheta}\bigl(m_t\bx_0+\sigma_t\sqrt{\tilde{S}_t}\,\bn,\;
    t,\;\log(\tilde{S}_t+10^{-6})\bigr) - \bn\Bigr\|_2^2 \Bigr].
  \label{eq:loss}
\end{equation}
The $10^{-6}$ is added to avoid numerical overflow when $\tilde{S}_t \to 0$. Since the factor $(\sigma_t^2\tilde{S}_t)^{-1}$ is strictly positive and conditionally decoupled from the network output, this unweighted objective is theoretically consistent with \eqref{eq:weighted_dsm_loss}. 

For the Gaussian case $\alpha(t)=2$ we set $S_t\equiv1$ exactly. For
$\alpha(t)<2$, sampling $S\sim\mathcal{A}_{\beta}$ with
$\beta=\alpha(t)/2\in(1/2,1)$ uses the
Kanter--Chambers--Mallows--Stuck representation (KCMS)
\citep{kanter1975stable,chambers1976method,devroye2009random}.
Algorithms~\ref{alg:ald_training}, \ref{alg:ald},
and~\ref{alg:ald_posterior} give the training, ALD-renoise, and
ALD-posterior procedures, respectively.

\begin{algorithm}[!t]
\caption{ALD Training}
\label{alg:ald_training}
\small
\begin{algorithmic}[1]
\REQUIRE Dataset $p_{\mathrm{data}}$, schedule $\alpha(\cdot)$, threshold $\tau$,
         clamp $S_{\max}$, learning rate $\gamma$
\REPEAT
  \STATE Sample $\bx_0 \sim p_{\mathrm{data}}$,\quad
         $t \sim \mathrm{Uniform}(0,1)$,\quad
         $\bn \sim \mathcal{N}(\mathbf{0},\mathbf{I})$
  \STATE $\alpha_t \leftarrow \alpha(t)$ via Eq.~\eqref{eq:alpha_schedule}
  \IF{$\alpha_t = 2$}
    \STATE $\tilde{S}_t \leftarrow 1$
    \hfill $\triangleright$ Gaussian refinement phase; no heavy-tail draw needed
  \ELSE
    \STATE Sample $S_t$ via KCMS  with $\beta = \alpha_t/2$
    \STATE $\tilde{S}_t \leftarrow \min(S_t,\,S_{\max})$
  \ENDIF
  \STATE $\bx_t \leftarrow m_t\bx_0 + \sigma_t\sqrt{\tilde{S}_t}\,\bn$
         \hfill (forward kernel, Eq.~\eqref{eq:marginal_forward})
  \STATE $\mathcal{L} \leftarrow
         \bigl\|\eps_{\btheta}\!\left(\bx_t,\;t,\;\log(\tilde{S}_t+10^{-6})\right)
         - \bn\bigr\|_2^2$
         \hfill (Eq.~\eqref{eq:loss})
  \STATE $\btheta \leftarrow \btheta - \gamma\,\nabla_{\btheta}\mathcal{L}$
\UNTIL{converged}
\RETURN $\btheta$
\end{algorithmic}
\end{algorithm}

\begin{algorithm}[!b]
\caption{ALD-Renoise Sampling}
\label{alg:ald}
\small
\begin{algorithmic}[1]
\REQUIRE Steps $N$, schedule $\alpha(\cdot)$, $S_{\max}$, $\tau$,
         trained network $\eps_{\btheta}$, data mean $\bar{\bx}_{\mathrm{data}}$
\STATE Draw $\tilde{S}_1$ via KCMS  with $\beta=\alpha(1)/2<1$, clamp to $S_{\max}$
\STATE Sample $\bx_1 \leftarrow \sqrt{\tilde{S}_1}\,\bn$, $\bn\sim\mathcal{N}(\mathbf{0},\mathbf{I})$
  \hfill $\triangleright$ exact draw from $p_1^{\mathrm{clamp}}$ (since $m_1=0$, $\sigma_1=1$)
\FOR{$i = N$ \textbf{downto} $1$}
  \STATE $t \leftarrow i/N$,\quad $\Delta t \leftarrow 1/N$
  \IF{$\alpha(t) = 2$}
    \STATE $\tilde{S}_t \leftarrow 1$ \hfill $\triangleright$ Gaussian refinement: force $S_t\equiv 1$
  \ENDIF
  \IF{$t=1$}
    \STATE $\hat{\bx}_0^{\mathrm R} \leftarrow \bar{\bx}_{\mathrm{data}}$
      \hfill (empirical estimate of $\mathbb{E}[\bx_0]$)
  \ELSE
    \STATE $\hat{\bx}_0^{\mathrm R} \leftarrow
      \dfrac{\bx_t - \sigma_t\sqrt{\tilde{S}_t}\,
      \eps_{\btheta}\!\left(\bx_t,\;t,\;\log(\tilde{S}_t+10^{-6})\right)}{m_t}$
      \hfill (Eq.~\eqref{eq:tweedie})
  \ENDIF
  \IF{$\alpha(t-\Delta t) = 2$}
    \STATE $\tilde{S}_{t-\Delta t} \leftarrow 1$
  \ELSE
    \STATE Sample $S$ via KCMS representation with $\beta=\alpha(t-\Delta t)/2$
    \STATE $\tilde{S}_{t-\Delta t} \leftarrow \min(S, S_{\max})$
  \ENDIF
  \STATE Sample $\bn \sim \mathcal{N}(\mathbf{0},\mathbf{I})$ independently of the past and $\tilde S_{t-\Delta t}$
  \STATE $\bx_{t-\Delta t} \leftarrow
    m_{t-\Delta t}\,\hat{\bx}_0^{\mathrm R}
    + \sigma_{t-\Delta t}\sqrt{\tilde S_{t-\Delta t}}\,\bn$
    \hfill 
\ENDFOR
\RETURN $\bx_0$
\end{algorithmic}
\end{algorithm}

\begin{algorithm}[!b]
\caption{ALD-Posterior Sampling }
\label{alg:ald_posterior}
\small
\begin{algorithmic}[1]
\REQUIRE Steps $N$, schedule $\alpha(\cdot)$, clamp $S_{\max}$,
         trained network $\eps_{\btheta}$, data mean $\bar{\bx}_{\mathrm{data}}$
\STATE $\Delta t\leftarrow 1/N$, $V_0\leftarrow0$
\FOR{$i=1$ \textbf{to} $N-1$}
  \STATE $t\leftarrow i/N$
  \IF{$\alpha(t)=2$}
    \STATE $\tilde S_t\leftarrow1$
  \ELSE
    \STATE Sample $S_t$ via KCMS with $\beta=\alpha(t)/2$;
           $\tilde S_t\leftarrow\min(S_t,S_{\max})$
  \ENDIF
  \STATE $V_t\leftarrow\dfrac{m_t^2}{m_{t-\Delta t}^2}V_{t-\Delta t}
    +\left(\sigma_t^2-\dfrac{m_t^2}{m_{t-\Delta t}^2}
      \sigma_{t-\Delta t}^2\right)\tilde S_t$
    \hfill (Eq.~\eqref{eq:ald_posterior_variance})
\ENDFOR
\STATE $t\leftarrow1-\Delta t$; sample
  $\bx_t\leftarrow m_t\bar{\bx}_{\mathrm{data}}+\sqrt{V_t}\,\bn$,
  $\bn\sim\mathcal N(\mathbf0,\mathbf I)$
\FOR{$i=N-1$ \textbf{downto} $1$}
  \STATE $t\leftarrow i/N$
  \STATE $\hat\bx_0^{\mathrm P}\leftarrow
    \dfrac{\bx_t-\sqrt{V_t}\,
      \eps_{\btheta}(\bx_t,t,\log(V_t/\sigma_t^2+10^{-6}))}{m_t}$
  \STATE In Eqs.~\eqref{eq:ald_posterior_mean}--
    \eqref{eq:ald_posterior_covariance}, replace $\bx_0$ by
    $\hat\bx_0^{\mathrm P}$
  \STATE Sample $\bx_{t-\Delta t}$ using the resulting Gaussian mean and
    covariance
\ENDFOR
\RETURN $\bx_0$
\end{algorithmic}
\end{algorithm}

\section{Numerical Experiments}
\label{sec:experiments}
We study distributional fidelity on synthetic targets,
training efficiency on heavy-tailed data, image generation on MNIST and
CelebA64, and downstream data augmentation on CIFAR-10LT.
Each experiment states its objective, dataset, protocol, and findings.

\subsection{Experimental Setup}
\label{sec:exp:setup}

\paragraph{Models and metrics.}
We compare ALD with DDPM \cite{ho2020ddpm} and DLPM
\cite{shariatian2025dlpm}. DDPM uses Gaussian corruption, DLPM a fixed
stability index, and ALD the schedule $\alpha(t)$ with clamped scale
$\tilde S_t$. We report sliced $W_2$ and Gaussian-kernel $\mathrm{MMD}^2$
as standard distributional metrics; lower is better. Their estimators,
bandwidth rule, sampling parameters, and reporting conventions are given in
Appendix~\ref{app:exp_details}.
For CelebA64 image generation, we additionally report Fr\'echet Inception
Distance (FID), Kernel Inception Distance (KID), and feature-space
precision, recall, density, and coverage.
Lower FID and KID and higher precision, recall, density, and coverage
are preferred. The evaluation protocol is specified in
Section~\ref{exp:celeba64}.

\paragraph{Sliced Hallucination Score (SHS).}
SHS is our diagnostic for spurious directional extremes among generated tail
samples. Let $\mathcal R=\{\bx_i\}_{i=1}^{n}$ and $\mathcal G$ be the real
and generated samples. Standardize each coordinate using $\mathcal R$, retain
the $h=\max(1,\lfloor d/2\rfloor)$ coordinates $\mathcal H$ with largest
empirical excess kurtosis, and let $\mathcal G_\xi$ be the generated samples
above the real-data radial quantile $\xi=0.95$ in this standardized subspace.
For $M=200$ random unit directions $\bv_\ell$, set
$q_\ell=Q_{0.99}(\{|\bv_\ell^\top\bx_i|\}_{i=1}^{n})$ and
\[
 A(\mathcal R,\mathcal G)=
 \frac{1}{M|\mathcal G_\xi|}
 \sum_{\by\in\mathcal G_\xi}\sum_{\ell=1}^{M}
 \mathbf1[|\bv_\ell^\top\by|>q_\ell]-0.01.
\]
We report $\mathrm{SHS}=A(\mathcal R,\mathcal G)/A(\mathcal R,\mathcal R')$
using an independent real set $\mathcal R'$; values closer to one indicate
better agreement with this real-versus-real baseline. The exact quantile,
standardization, and validity conventions are in
Appendix~\ref{app:exp_details}.

\subsection{Experiment 1: Generation Accuracy on 2D Synthetic Benchmarks}
\label{exp:synthetic}

\paragraph{Datasets and protocol.}
We use a heavy-tailed cross (HTC), a Swiss Roll manifold, a forty-component
Gaussian mixture (MoG40), and a Holtsmark distribution as the four main 2D
benchmarks, together with an imbalanced nine-mode Gaussian mixture for
mode-mass recovery. For the four main benchmarks, each comparison uses five
seeds and matched reference samples. ALD-renoise uses dataset-specific
sampling configurations. Sampling configurations are selected on validation
streams and frozen before final evaluation. Dataset constructions,
architectures, training budgets, clipping bounds, and ALD-posterior
selection details are reported in Appendix~\ref{app:exp_details}.

\begin{table}[htbp]
\centering
\caption{2D benchmarks (mean $\pm$ sample standard deviation over
five seeds). Lower is better for $W_2$ and $\mathrm{MMD}^2$; SHS is ranked
by distance to one. Swiss Roll SHS uses the two valid seeds shared by all
methods. Bold marks the best mean. Full
protocol details are in Appendix~\ref{app:exp_details}.}
\label{tab:quantitative}
\small
\setlength{\tabcolsep}{4pt}
\resizebox{\linewidth}{!}{\begin{tabular}{llccc}
\toprule
Dataset & Model & $W_2$ ($\downarrow$) & $\mathrm{MMD}^2$ ($\downarrow$) & SHS ($\approx1$) \\
\midrule
\multirow{4}{*}{Heavy-Tailed Cross}
 & DDPM & $0.0833\pm0.0195$ & $0.000852\pm0.001177$ & $0.8857\pm0.2411$ \\
 & DLPM & $0.1772\pm0.0529$ & $0.005861\pm0.008071$ & $0.5817\pm0.2223$ \\
 & ALD-renoise & $\mathbf{0.0830\pm0.0374}$ & $\mathbf{0.000384\pm0.000361}$ & $1.1131\pm0.3757$ \\
 & ALD-posterior & $0.0884\pm0.0113$ & $0.000435\pm0.000763$ & $\mathbf{0.9398\pm0.2344}$ \\
\midrule
\multirow{4}{*}{Swiss Roll}
 & DDPM & $0.0268\pm0.0081$ & $0.000267\pm0.000489$ & $\mathbf{1.2473\pm0.3656}$ \\
 & DLPM & $0.0971\pm0.0207$ & $0.003356\pm0.002302$ & $0.6516\pm0.3785$ \\
 & ALD-renoise & $0.0280\pm0.0030$ & $\mathbf{0.000000\pm0.000000}$ & $1.3951\pm0.5186$ \\
 & ALD-posterior & $\mathbf{0.0266\pm0.0055}$ & $0.000235\pm0.000397$ & $1.3277\pm0.5026$ \\
\midrule
\multirow{4}{*}{MoG40}
 & DDPM & $\mathbf{0.0923\pm0.0140}$ & $0.000750\pm0.001367$ & $1.1732\pm0.2617$ \\
 & DLPM & $0.2342\pm0.0514$ & $0.001291\pm0.001702$ & $\mathbf{0.9519\pm0.5407}$ \\
 & ALD-renoise & $0.1194\pm0.0101$ & $\mathbf{0.000473\pm0.000542}$ & $1.2651\pm0.5407$ \\
 & ALD-posterior & $0.0960\pm0.0270$ & $0.000478\pm0.000840$ & $1.1318\pm0.2214$ \\
\midrule
\multirow{4}{*}{Holtsmark Field}
 & DDPM & $0.4919\pm0.0836$ & $0.000617\pm0.000826$ & $0.6132\pm0.3894$ \\
 & DLPM & $1.4924\pm0.1378$ & $0.030361\pm0.005575$ & $-0.0397\pm0.0795$ \\
 & ALD-renoise & $0.6195\pm0.0443$ & $0.009244\pm0.000738$ & $0.3707\pm0.3524$ \\
 & ALD-posterior & $\mathbf{0.4701\pm0.0461}$ & $\mathbf{0.000494\pm0.000489}$ & $\mathbf{1.0058\pm0.4758}$ \\
\bottomrule
\end{tabular}}
\end{table}

\begin{figure}[htbp]
\centering
\includegraphics[width=\linewidth]{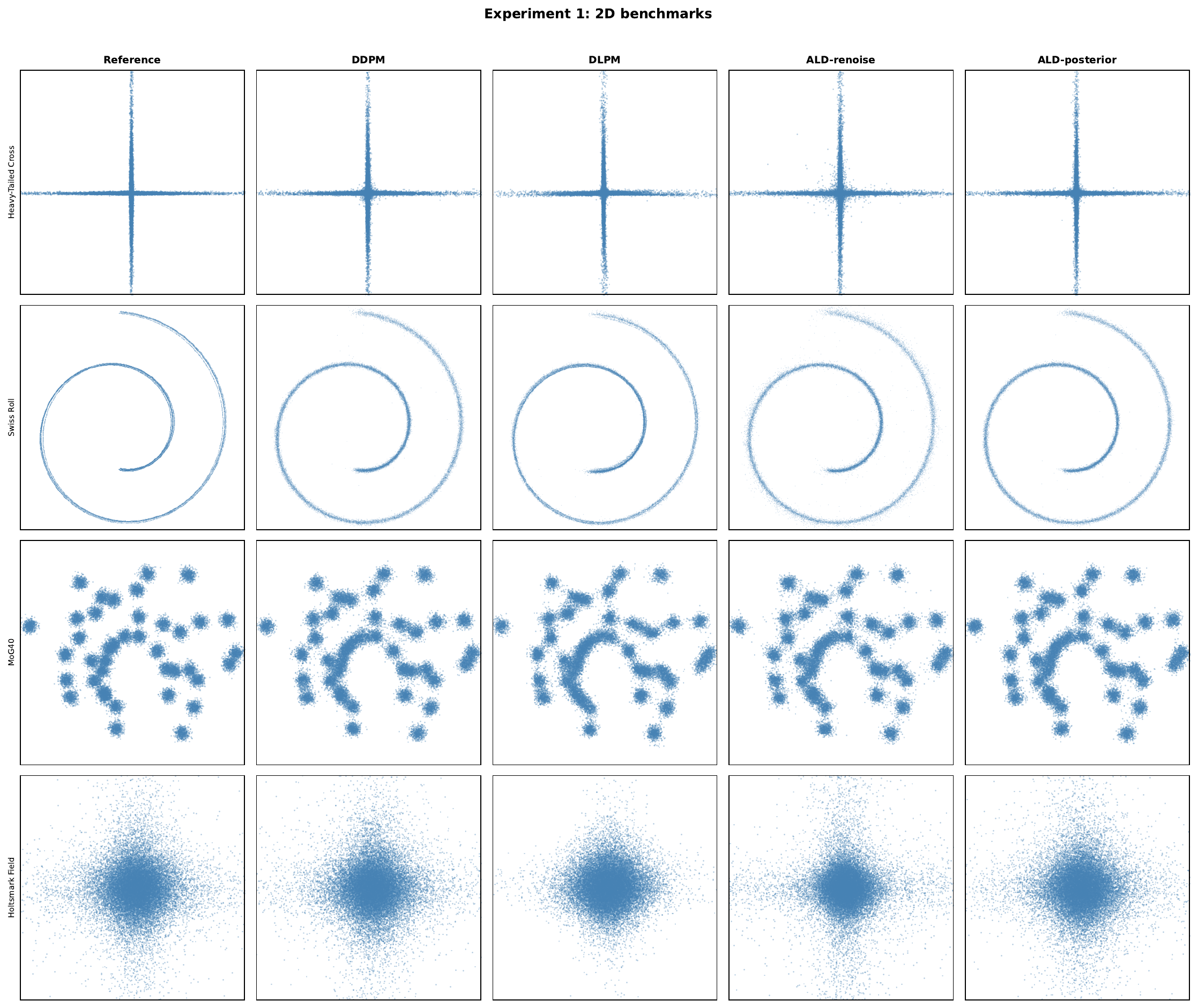}
\caption{{Representative samples on the four main 2D
benchmarks. Rows correspond to HTC, Swiss Roll, MoG40, and Holtsmark;
columns show the reference data, DDPM, DLPM, ALD-renoise, and ALD-posterior.
The quantitative metrics are reported separately in
Table~\ref{tab:quantitative}.}}
\label{fig:exp1_qualitative}
\end{figure}

\paragraph{Results and discussion.}
The 2D results show that ALD achieves competitive generation quality
on light-tailed and manifold-structured data, with ALD-posterior
providing stronger performance on heavy-tailed data.
On the light-tailed MoG40 benchmark and the manifold-structured HTC
and Swiss Roll benchmarks, ALD-renoise and ALD-posterior achieve
generation quality comparable to DDPM and better than DLPM
(Figure~\ref{fig:exp1_qualitative}).
This behavior is consistent with ALD's Gaussian refinement phase,
which helps recover local geometry after heavy-tailed exploration.
On the heavy-tailed Holtsmark benchmark, ALD-posterior performs best
among all compared methods (Table~\ref{tab:quantitative}).
This result supports combining L\'{e}vy exploration with the
variance-controlled posterior transition to capture global tail structure.

\paragraph{Additional dataset: imbalanced nine-mode Gaussian mixture.}
Class imbalance can produce long-tailed class-frequency distributions
\citep{cui2019classbalanced}.
We therefore examine an imbalanced nine-mode Gaussian mixture with
a $30{:}1$ largest-to-smallest mass ratio.
Following the DLPM implementation \citep{shariatian2025dlpm},
we report PRD precision, recall, and their harmonic mean
$F^{\rm pr}_1$ to summarize distributional fidelity and coverage.
The mixture weights, model settings, and exact PRD computation are
given in Appendix~\ref{app:exp_details}.

\begin{table}[H]
\centering
\caption{Additional imbalanced nine-mode benchmark (mean $\pm$ std over
three seeds; higher is better).}
\label{tab:exp1_9gaussian}
\begin{tabular}{lccc}
\toprule
Model & $P_{\rm PRD}$ ($\uparrow$) & $R_{\rm PRD}$ ($\uparrow$)
 & $F^{\rm pr}_1$ ($\uparrow$) \\
\midrule
DDPM & $0.9307\pm0.0177$ & $0.9805\pm0.0039$ & $0.9549\pm0.0104$ \\
DLPM & $0.9600\pm0.0064$ & $0.9835\pm0.0014$ & $0.9716\pm0.0039$ \\
DLPM5 & $0.9605\pm0.0031$ & $0.9820\pm0.0031$ & $0.9711\pm0.0011$ \\
ALD-renoise & $0.9897\pm0.0006$ & $0.9863\pm0.0026$
 & $0.9880\pm0.0016$ \\
ALD-posterior & $\mathbf{0.9909\pm0.0012}$ & $\mathbf{0.9935\pm0.0004}$
 & $\mathbf{0.9922\pm0.0004}$ \\
\bottomrule
\end{tabular}
\end{table}

\begin{figure}[H]
\centering
\includegraphics[width=\linewidth]{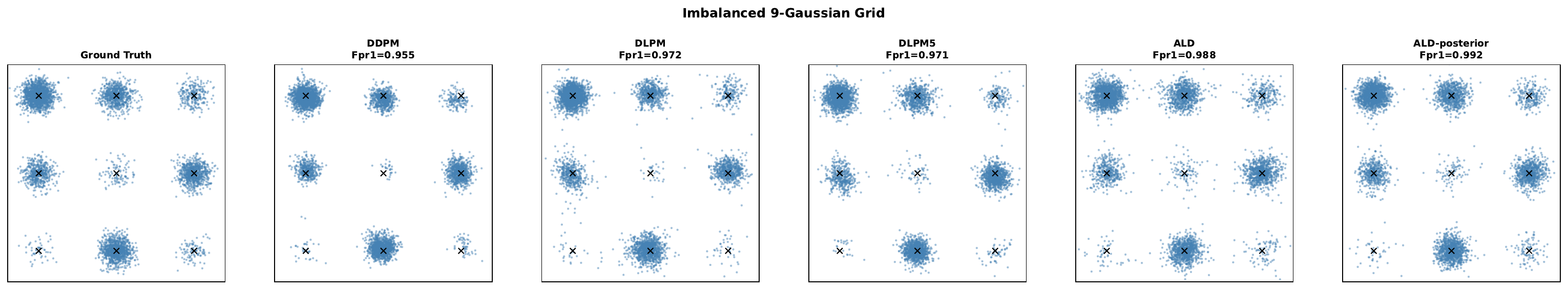}
\caption{{Additional nine-mode benchmark under unequal
mixture weights. }}
\label{fig:exp1_9gaussian}
\end{figure}

\paragraph{Additional result: recovery under mode imbalance.}
Both ALD-renoise and ALD-posterior retain rare modes and local
cluster structure under a $30{:}1$ mass imbalance.
ALD-renoise performs better than DDPM and both DLPM variants,
and is surpassed only by ALD-posterior
(Table~\ref{tab:exp1_9gaussian}).

\FloatBarrier

\subsection{Experiment 2: Training Efficiency on Heavy-Tailed Data}
\label{exp:holtsmark_robustness}

\paragraph{Dataset and protocol.}
We use the Holtsmark distribution and vary network width and optimizer
budget.
The complete sweep grids, data sizes, and optimization and sampling
settings are given in Appendix~\ref{app:exp_details}.

\begin{figure}[htbp]
\centering
\includegraphics[width=\linewidth]{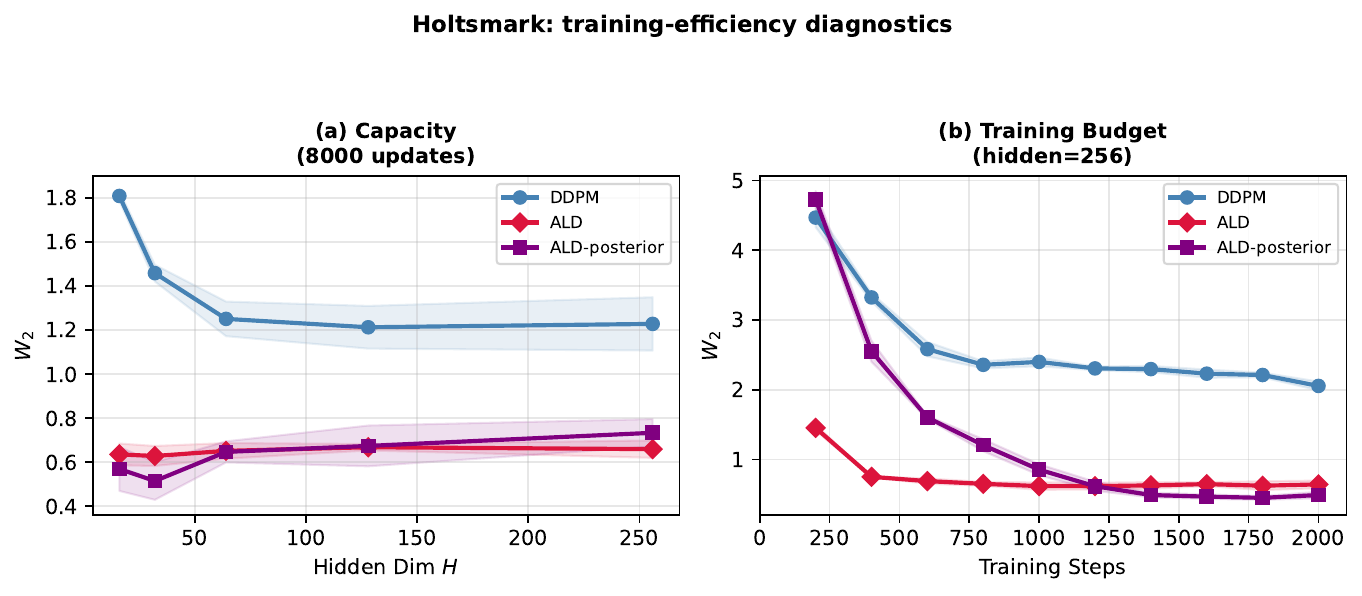}
\caption{Training-efficiency diagnostics on clipped Holtsmark
(mean $\pm$ std over three seeds): $W_2$ versus
(a) hidden width and (b) optimizer updates.
ALD-renoise and ALD-posterior share checkpoints and sampling
parameters; lower is better.
Full settings are in Appendix~\ref{app:exp_details}.}
\label{fig:exp2_holtsmark}
\end{figure}

\paragraph{Results and discussion.}
ALD-renoise performs better than ALD-posterior at smaller training
budgets.
With more training updates, ALD-posterior becomes more effective
and achieves better generation quality.
Across the tested network widths, both ALD-renoise and ALD-posterior
outperform DDPM, indicating that their advantage does not rely on
using a larger network.
The observed training-efficiency advantage of the ALD training
formulation is consistent with the benefit of heavy-tailed exploration.

\FloatBarrier

\subsection{Experiment 3:    Extension to MNIST}
\label{exp:mnist}
\paragraph{Dataset and protocol.}
We use $28\times28$ MNIST images normalized to $[-1,1]$ and report
classwise kNN coverage.
Configurations are selected on validation data and fixed before evaluation.
Each method uses one training seed and two sampling seeds.
Generated images are grouped by two frozen MNIST classifiers,
and the resulting diagnostics are averaged.
Higher coverage is better; all training, sampling, classifier,
and metric settings are given in Appendix~\ref{app:exp_details}.

\begin{table}[htbp]
\centering
\small
\setlength{\tabcolsep}{3pt}
\caption{Classwise MNIST kNN coverage (\%). Bold marks the highest coverage
in each digit column, including ties. Values average two classifier diagnostics
and two sampling seeds from one generator training seed; full settings are in
Appendix~\ref{app:exp_details}.}
\label{tab:mnist_classwise}
\begin{tabular}{l*{10}{r}}
\toprule
Method / Digit & 0 & 1 & 2 & 3 & 4 & 5 & 6 & 7 & 8 & 9 \\
\midrule
DDPM & 72.12 & 85.93 & 72.55 & 67.21 & 67.78 & 67.17 & 75.20 & 73.48 & 62.15 & 74.89 \\
DLPM & 86.83 & 91.48 & 88.42 & 85.76 & \textbf{88.51} & 86.48 & 86.25 & 87.79 & 81.40 & \textbf{89.37} \\
ALD-renoise & 74.23 & 85.56 & 66.95 & 60.13 & 62.43 & 64.06 & 70.96 & 70.77 & 61.35 & 74.02 \\
ALD-posterior & \textbf{89.20} & \textbf{93.43} & \textbf{91.53} & \textbf{87.76} & \textbf{88.51} & \textbf{88.79} & \textbf{89.67} & \textbf{90.40} & \textbf{84.38} & 89.26 \\
\bottomrule
\end{tabular}
\end{table}

\paragraph{Results and discussion.}
ALD-posterior achieves higher coverage than ALD-renoise on all ten digits.
It obtains the best or tied coverage on nine of ten digits,
while DLPM is best on digit 9.

\FloatBarrier

\subsection{Experiment 4: Application to CIFAR-10LT}
\label{exp:cifar10_augmentation}
\paragraph{Dataset and protocol.}
We use the 100{:}1 CIFAR-10LT split to train class-conditional generators under
matched architectures, training budgets, and sampling budgets.
Each model produces a synthetic set that is pseudo-labeled by a common
ResNet-18 teacher and combined with balanced real CIFAR-10 data to train
a new classifier.
We evaluate accuracy and one-vs-rest classwise F$_1$ on the balanced test set.
Exact class counts, data sizes, and generator, sampler, teacher, and classifier
hyperparameters are given in Appendix~\ref{app:exp_details};
each augmentation arm has one classifier run.

\begin{table}[htbp]
\centering
\small
\caption{CIFAR-10 test accuracy (\%). Bold marks the best result among
the three synthetic-augmentation arms; the teacher model is shown as a
real-data-only reference.}
\label{tab:cifar10_aug_accuracy}
\begin{tabular}{lc}
\toprule
Training data & Accuracy \\
\midrule
Teacher model & 93.43 \\
DDPM + Real & 91.61 \\
DLPM + Real & 91.42 \\
ALD-posterior + Real & \textbf{91.72} \\
\bottomrule
\end{tabular}
\end{table}

\begin{table}[htbp]
\centering
\scriptsize
\setlength{\tabcolsep}{2.5pt}
\caption{Classwise CIFAR-10 F$_1$ (\%). Bold marks the best result in
each column among the three synthetic-augmentation arms; the teacher
model is excluded from this comparison.}
\label{tab:cifar10_aug_f1}
\resizebox{\linewidth}{!}{%
\begin{tabular}{l*{11}{r}}
\toprule
Method / Class & Airplane & Auto. & Bird & Cat & Deer & Dog & Frog & Horse & Ship & Truck & Macro \\
\midrule
Teacher model & 94.11 & 96.59 & 91.32 & 85.97 & 93.78 & 90.14 & 95.24 & 95.65 & 95.96 & 95.63 & 93.44 \\
DDPM + Real & 91.37 & 96.11 & 88.53 & 82.28 & \textbf{92.37} & 86.59 & \textbf{94.67} & \textbf{94.71} & 94.68 & 94.77 & 91.61 \\
DLPM + Real & \textbf{92.55} & 96.17 & 88.07 & 81.80 & 91.55 & 86.15 & 93.97 & 93.49 & \textbf{95.39} & 94.98 & 91.41 \\
ALD-posterior + Real & 92.14 & \textbf{96.57} & \textbf{88.69} & \textbf{82.50} & 91.83 & \textbf{87.03} & 93.96 & 94.35 & 94.85 & \textbf{95.18} & \textbf{91.71} \\
\bottomrule
\end{tabular}%
}
\end{table}

\paragraph{Results.}
 The ALD-posterior samples provide the strongest augmentation arm overall.
This suggests that their broader support remains useful downstream.

\subsection{Experiment 5: Performance on CelebA64}
\label{exp:celeba64}

\paragraph{Dataset and protocol.}
We compare unconditional DDPM, DLPM, and ALD-posterior generation on CelebA64
under matched network architectures, training and sampling budgets,
and evaluation protocols.
Detailed experimental settings and validation-only sampler selection
are given in Appendix~\ref{app:celeba64_sampling}.

\begin{table*}[t]
\centering
\small
\setlength{\tabcolsep}{5pt}
\caption{Formal CelebA64 generation results (single training and sampling
seed; 50,000 generated images and 1,000 NFE per method). Lower is better for
FID and KID, higher for the four PRDC metrics. KID values and their subset
standard deviations are multiplied by $10^3$. Bold marks the best reported
value in each column, not statistical significance.}
\label{tab:celeba64_formal}
\begin{tabular}{lcccccc}
\toprule
Model & FID$\downarrow$ & KID$\times10^3\downarrow$
& Precision$\uparrow$ & Recall$\uparrow$ & Density$\uparrow$
& Coverage$\uparrow$ \\
\midrule
DDPM & $2.5763\pm0.0197$ & $1.7926\pm0.4276$ & 0.7607 & 0.5819 & 1.2721 & 0.9680 \\
DLPM & $2.9660\pm0.0221$ & $1.9828\pm0.3872$ & \textbf{0.7681} & 0.5687
& \textbf{1.3417} & \textbf{0.9697} \\
ALD-posterior & $\mathbf{2.1090\pm0.0155}$ & $\mathbf{1.1419\pm0.2892}$ & 0.7536
& \textbf{0.5940} & 1.2417 & 0.9683 \\
\bottomrule
\end{tabular}
\end{table*}

\begin{figure}[htbp]
\centering
\includegraphics[width=0.7\linewidth]{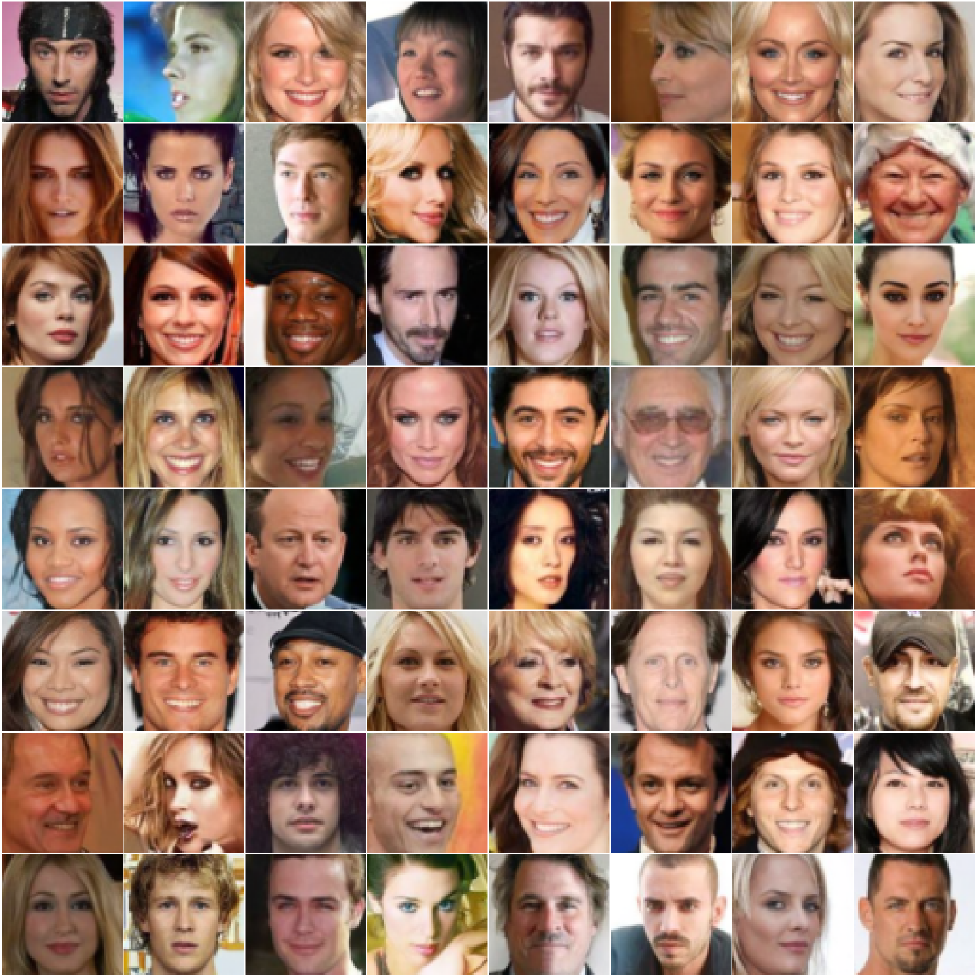}
\caption{Uncurated ALD-posterior CelebA64 cases.}
\label{fig:celeba64_ald_cases}
\end{figure}

\paragraph{Results and discussion.}
At the frozen settings, ALD-posterior obtains the lowest FID (2.1090) and KID
($1.1419\times10^{-3}$) and the highest recall (0.5940). DLPM instead has
the highest precision (0.7681), density (1.3417), and coverage (0.9697),
while its FID and KID are higher and its recall is lower.
These results extend the evaluation of ALD-posterior to natural-image generation,
with improved FID, KID, and recall over both baselines in this run.
The representative ALD-posterior samples are shown in
Figure~\ref{fig:celeba64_ald_cases}.

\paragraph{Runtime.}
On four V100 GPUs, training elapsed 5d 06h 22m for DDPM,
5d 08h 04m for DLPM, and 4d 13h 31m for ALD-posterior.
Total generation time was 7h 23m for DDPM,
8h 22m 40s for DLPM, and 8h 18m for ALD-posterior.

\section{Discussion \& Conclusion}
\label{sec:discussion}
\label{sec:conclusion}

\paragraph{Training budget and prediction accuracy.}
ALD-renoise and ALD-posterior share the same noise-prediction
training formulation.
In the Holtsmark experiments, ALD-renoise performs better at smaller
training budgets, whereas ALD-posterior becomes more effective
after more training updates.
These results suggest that the preferred sampler depends on
prediction accuracy.
The controlled prediction-output perturbation experiment further
supports this tradeoff: ALD-renoise is less sensitive to perturbation added to the network's
predictions, while ALD-posterior achieves better generation quality
without this added noise, as observed in the Holtsmark experiment
(Appendix~\ref{app:prediction_perturbation}).

\paragraph{Mathematical interpretation of ALD-posterior.}
The forward perturbation in Eq.~\eqref{eq:marginal_forward}
specifies the distribution at each time but leaves the dependence
between times unspecified.
Whether the process is Markovian cannot be determined from these
marginal distributions alone.
ALD-posterior introduces an explicit Markov forward process through
Eq.~\eqref{eq:ald_posterior_forward}, with accumulated conditional
variance $V_t$ given by Eq.~\eqref{eq:ald_posterior_variance}.
This construction replaces the original conditional noise variance
$\sigma_t^2S_t$ by $V_t$ and generally changes the intermediate
marginal distributions.
It nevertheless retains the same initial data distribution,
since $m_0=1$ and $V_0=0$.

For this new forward process, Gaussian conditioning gives an exact
adjacent posterior when the clean sample $\bx_0$ and the relevant
variance variables are known.
During generation, $\bx_0$ is replaced by the denoiser
$\hat\bx_0^{\mathrm P}$.
The resulting Gaussian approximation omits the contribution of
clean-data posterior uncertainty to the reverse covariance.
ALD-posterior is thus a variance-controlled reverse sampler for
an alternative forward process, whose intermediate marginals
differ from those used by ALD-renoise.

\paragraph{Conclusion.}
We proposed Annealed L\'{e}vy Diffusion (ALD), which combines
heavy-tailed exploration with Gaussian refinement while retaining
conditionally Gaussian noise-prediction training.
Experiments on synthetic and image benchmarks show competitive
generation quality, rare-mode coverage, and training efficiency.

\appendix

\section{Experimental Metrics and Protocol Details}
\label{app:exp_details}

{
\subsection{Metrics and Reporting Conventions}

Let $\mathcal R=\{\bx_i\}_{i=1}^{n}\subset\R^d$ and
$\mathcal G=\{\by_j\}_{j=1}^{m}\subset\R^d$ denote real and generated
samples. For $A=\{a_1,\ldots,a_N\}\subset\R$, write
$a_{(1)}\le\cdots\le a_{(N)}$ and use the nearest-rank empirical quantile
$Q_q(A)=a_{(\lceil qN\rceil)}$. The indicator $\mathbf1[E]$ is one when
$E$ holds and zero otherwise. Unless stated otherwise, mean $\pm$ std uses
the population standard deviation across random seeds; Table~\ref{tab:quantitative}
uses the sample standard deviation.

\paragraph{Sliced 2-Wasserstein distance.}
The reported $W_2$ is a sliced distance rather than the full
$d$-dimensional optimal-transport distance. For equal-sized sample sets, we
draw $M=200$ unit directions $\bv_\ell$ uniformly on $\mathbb S^{d-1}$ and
compute
\[
 \widehat W_{2,\mathrm{slice}}(\mathcal R,\mathcal G)
 =\left[\frac{1}{Mn}\sum_{\ell=1}^{M}\sum_{i=1}^{n}
 \left(a_{\ell,(i)}-b_{\ell,(i)}\right)^2\right]^{1/2},
\]
where $a_{\ell,(i)}$ and $b_{\ell,(i)}$ are the $i$th order statistics of
$\{\bv_\ell^\top\bx_i\}_{i=1}^{n}$ and
$\{\bv_\ell^\top\by_i\}_{i=1}^{n}$.

\paragraph{Maximum mean discrepancy.}
We use at most 1{,}000 points from each set and the Gaussian kernel
$k_\ell(\bx,\by)=\exp[-\|\bx-\by\|_2^2/(2\ell^2)]$. Its bandwidth obeys
$\ell^2=\tfrac12Q_{0.5}(D_+)+10^{-8}$, where
$D_+=\{\|\bx-\by\|_2^2:\bx,\by\in\mathcal R\cup\mathcal G,\ \bx\ne\by\}$.
The displayed value is
\[
 \max\!\left\{0,\,
 \frac{\sum_{i\ne i'}k_\ell(\bx_i,\bx_{i'})}{n(n-1)}
 +\frac{\sum_{j\ne j'}k_\ell(\by_j,\by_{j'})}{m(m-1)}
 -\frac{2\sum_{i,j}k_\ell(\bx_i,\by_j)}{nm}\right\}.
\]
It is the unbiased MMD-squared estimate before truncation, clipped below at
zero for reporting.

\paragraph{SHS implementation details.}
Write $(\bx_i)_j$ for coordinate $j$ of real sample $\bx_i$, and set
\[
 \mu_j=\frac1n\sum_i(\bx_i)_j,\qquad
 s_j^2=\frac1n\sum_i((\bx_i)_j-\mu_j)^2,\qquad
 z_j(\bx)=\frac{(\bx)_j-\mu_j}{s_j+10^{-8}}.
\]
The selected coordinate set $\mathcal H$ contains the
$h=\max(1,\lfloor d/2\rfloor)$ largest empirical Fisher excess kurtoses
\[
 \kappa_j=
 \frac{n^{-1}\sum_i((\bx_i)_j-\mu_j)^4}
      {(n^{-1}\sum_i((\bx_i)_j-\mu_j)^2)^2}-3.
\]
All evaluated 2D coordinates have positive empirical variance. With
$\xi=0.95$,
\[
 r_\xi=Q_\xi(\{\|z_{\mathcal H}(\bx_i)\|_2\}_{i=1}^{n}),\qquad
 \mathcal G_\xi=\{\by\in\mathcal G:\|z_{\mathcal H}(\by)\|_2>r_\xi\}.
\]
The directional projections in $A(\mathcal R,\mathcal G)$ use the original
coordinates, while only the tail selector uses standardized coordinates.
Each method uses the same reference pair and projection seed. SHS is undefined
when $\mathcal G_\xi$ is empty or the real-versus-real denominator is at most
$10^{-10}$. It is conditional on the selected tail and does not estimate total
tail mass or a power-law exponent.

\paragraph{PRD summaries.}
For the imbalanced mixture, we retain the DLPM implementation's
precision--recall distribution summary. For each of ten joint clusterings into
100 bins, let $r_k$ and $g_k$ be reference and generated frequencies. For 201
slopes $a=\tan\phi$, with $\phi$ evenly spaced between $10^{-10}$ and
$\pi/2-10^{-10}$, define
\[
 P(a)=\sum_k\min\{a r_k,g_k\},\qquad R(a)=P(a)/a,
\]
and average the curves across clusterings. With
$F_\beta(P,R)=(1+\beta^2)PR/(\beta^2P+R)$, we report
\[
 P_{\rm PRD}=\max_aF_8(P(a),R(a)),\qquad
 R_{\rm PRD}=\max_aF_{1/8}(P(a),R(a)),
\]
and $F^{\rm pr}_1=2P_{\rm PRD}R_{\rm PRD}/(P_{\rm PRD}+R_{\rm PRD})$.
Zero-over-zero ratios are set to zero, with $10^{-10}$ denominator
regularization.

\paragraph{MNIST coverage.}
Each generated image defines a ball in the frozen
classifier's 128-dimensional feature space, with radius equal to its distance
to its third nearest generated neighbor. Classwise coverage is the fraction
of real images of a digit contained in the union of these balls. It is not the
generated class frequency.

\subsection{Two-Dimensional Benchmarks}

\paragraph{Dataset construction.}
HTC is an equal mixture of $(U,Z)$ and $(Z,U)$, where
$U\sim t_{1.5}$ and $Z\sim\mathcal N(0,0.05^2)$ are independent.
Each coordinate is clipped to $[-20,20]$ and divided by four,
giving samples in $[-5,5]^2$.
Swiss Roll uses $(r\cos r,r\sin r)$ with
$r\sim\operatorname{Unif}[3\pi/2,9\pi/2]$ and no added noise.
Each sampled set is centered and divided by its largest coordinate
standard deviation, then multiplied by 0.9.
MoG40 is an equally weighted mixture of forty isotropic Gaussians
with covariance $0.2^2\mathbf I_2$; centers are drawn from
$\mathcal N(\mathbf0,9\mathbf I_2)$ using seed 2024. Holtsmark Field uses two
independent symmetric $3/2$-stable coordinates with characteristic function
$\mathbb E[e^{iuX}]=e^{-|u|^{3/2}}$, each clipped to $[-20,20]$. It is an
independent-coordinate surrogate, not an isotropic physical field projection.

\paragraph{Common protocol.}
Each main dataset uses 32{,}000 training points per seed and seeds 0--4;
evaluation compares 5{,}000 generated with 5{,}000 reference points shared
across methods. ALD and DDPM use residual MLPs with Mish activations; ALD also
embeds $\log(\tilde S_t+10^{-6})$. DLPM uses time conditioning, LayerNorm, and
SiLU. All runs use batch size 512. Table~\ref{tab:exp_params} gives the
remaining training settings.

\begin{table}[htbp]
\centering

\small
\caption{Training settings for the main 2D benchmarks.}
\label{tab:exp_params}
\begin{tabular}{lccc}
\toprule
Dataset & Hidden dimension & Training steps & Learning rate \\
\midrule
Heavy-Tailed Cross & 128 & 15{,}000 & $5\times10^{-4}$ \\
Swiss Roll & 128 & 15{,}000 & $10^{-3}$ \\
MoG40 & 256 & 20{,}000 & $5\times10^{-4}$ \\
Holtsmark Field & 256 & 1{,}000 & $5\times10^{-4}$ \\
\bottomrule
\end{tabular}
\end{table}

DDPM uses fixed-small ancestral posterior variance on a discretized
trigonometric schedule. Coordinate-wise reconstruction bounds are 5, 3, 10,
and 14 in dataset order. DDPM and the matched DLPM rerun apply the bound to
the clean reconstruction at every reverse step; ALD-posterior applies its
bound to the final output. For DLPM we fix
$\alpha=(\alpha_{\min}^{\mathrm{ALD}}+2)/2$, giving 1.95, 1.85, 1.95, and
1.55. Its positive-stable variable $A$ uses the convention $A=2S$ relative
to ALD's Gaussian-normalized scale, so the training and sampling caps are
$A_{\max}=3,6,3,150$, corresponding to ALD caps $S_{\max}=1.5,3,1.5,75$.
DLPM uses 199 network evaluations on the first three datasets and 74 on
Holtsmark, compared with reverse grids of 200, 200, 200, and 75 points. No
DLPM hyperparameter search is performed.

\paragraph{ALD-posterior selection.}
For each dataset, we jointly search $\alpha_{\min}$, $\tau$, $S_{\max}$, and
the sampling grid without retraining. Candidates use separate validation
streams for checkpoint seeds 0--2 and are ranked by mean validation $W_2$.
The frozen $(\alpha_{\min},\tau,S_{\max})$ values are
$(1.95,0.925,10)$ for HTC, $(1.7,0,3)$ for Swiss Roll,
$(1.6,0.875,1)$ for MoG40, and $(1.87,0,56.25)$ for Holtsmark.
The corresponding sampling budgets are 400, 200, 400, and 50 NFE.
Baselines and
the original ALD receive no corresponding extra search budget. The final
reference sets and checkpoints were examined in earlier comparisons, so this
is not an untouched confirmatory study.

\paragraph{Imbalanced nine-mode mixture.}
The centers form a $3\times3$ grid on $\{0,1,2\}^2$, with covariance
$0.1^2\mathbf I_2$ and weights
\[
 (0.01,0.10,0.30,0.20,0.02,0.15,0.02,0.15,0.05).
\]
The benchmark uses 32{,}000 training points; 15{,}000 real and generated
evaluation points; three seeds; MLP width 64; batch size 1{,}024; 10{,}000
updates; learning rate $5\times10^{-3}$; and 100 reverse steps. DLPM5 uses
five Monte Carlo samples. ALD-posterior reuses the three ALD checkpoints,
annealing parameters, reconstruction clamp, and 100 network evaluations.

\paragraph{Holtsmark efficiency sweeps.}
The training set has 32{,}000 points; each evaluation uses 5{,}000 generated
and reference points and three seeds. Batch size is 512, learning rate is
$5\times10^{-4}$, EMA decay is 0.99, and sampling uses 75 reverse steps with
reconstruction clipping at 14. ALD uses
$(\alpha_{\min},\tau,S_{\max})=(1.1,0,75)$, and ALD-posterior shares these
settings. We vary hidden width
$H\in\{16,32,64,128,256\}$ at 8{,}000 updates, optimizer budget
$K\in\{200,400,\ldots,2000\}$ at $H=256$.

\subsection{Prediction-Output Perturbation}
\label{app:prediction_perturbation}

We reuse the three ALD checkpoints with $H=256$ and
$K=1{,}000$ from the Holtsmark efficiency study and perturb every network
output as
$\eps_{\btheta}+\sigma_{\mathrm{noise}}\bz$, where
$\bz\sim\mathcal N(\mathbf0,\mathbf I_2)$ and
$\sigma_{\mathrm{noise}}\in\{0,0.05,0.1,0.25,0.5,0.75,1\}$.
Each checkpoint is sampled three times, giving nine runs per condition; every
run compares 5{,}000 generated and 5{,}000 reference samples. All other
settings are shared. ALD-renoise makes 74 network calls because the endpoint
$t=1$ uses the empirical data mean, whereas ALD-posterior makes 75.

\begin{figure}[htbp]
\centering
\includegraphics[width=\linewidth]{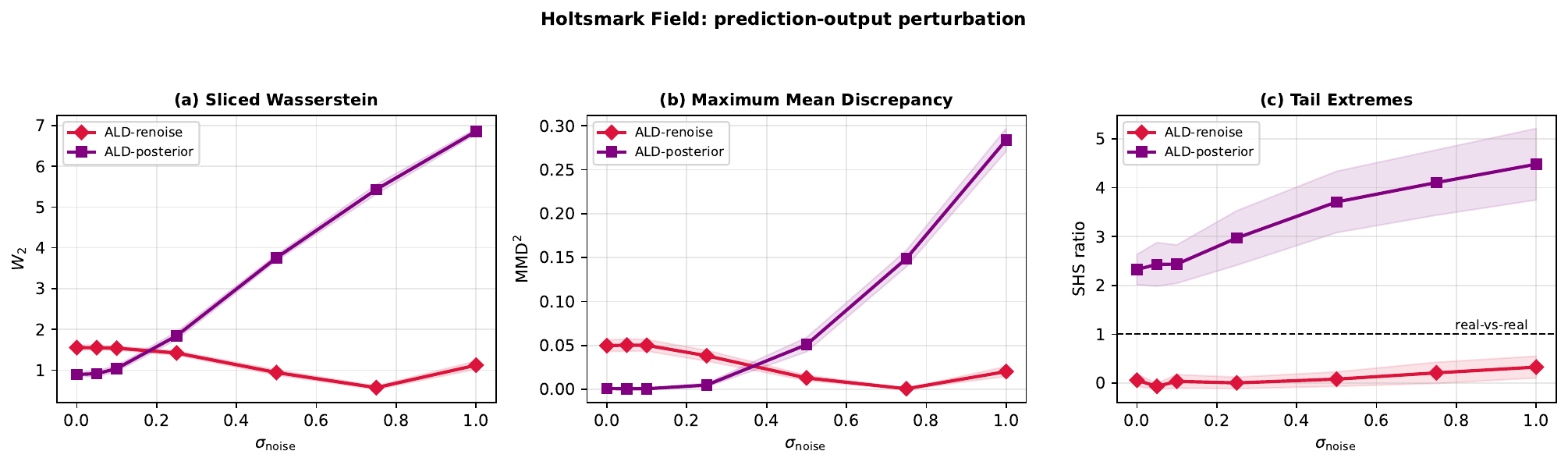}
\caption{Prediction-output perturbation on Holtsmark
(mean $\pm$ sample standard deviation over nine runs). Lower is better for
$W_2$ and $\mathrm{MMD}^2$; SHS is ranked by distance to one. Both samplers
reuse the same ALD checkpoints and differ only in the reverse transition.}
\label{fig:prediction_perturbation}
\end{figure}

Without perturbation, ALD-posterior has the lower $W_2$ and
$\mathrm{MMD}^2$. Small perturbations leave ALD-renoise nearly unchanged but
already increase the posterior $W_2$; by
$\sigma_{\mathrm{noise}}=0.25$, ALD-renoise has the lower $W_2$, and the gap
widens thereafter. The improvement of ALD-renoise at intermediate noise
levels should not be interpreted as monotone robustness: the added prediction
noise expands its initially concentrated samples, while its $W_2$ and
$\mathrm{MMD}^2$ worsen again at $\sigma_{\mathrm{noise}}=1$. SHS further
shows different tail biases---ALD-renoise underproduces directional extremes,
whereas ALD-posterior increasingly overproduces them. The experiment therefore
supports an accuracy--sensitivity trade-off between the two reverse rules,
not uniform superiority of either rule.

\subsection{MNIST Protocol}

MNIST has 60{,}000 $28\times28$ grayscale training images normalized to
$[-1,1]$. We train for 6{,}000 updates with U-Net base width 32, batch size
128, learning rate $2\times10^{-4}$, and training seed 42. Two sets of 5{,}000
samples use seeds 161803 and 271828 and share 5{,}000 held-out test images.
Both ALD variants use $(\alpha_{\min},\tau,S_{\max})=(1.5,0.15,8)$; DLPM uses
fixed $\alpha=1.8$ and scale cap 20. DDPM, DLPM, ALD-renoise, and
ALD-posterior use 10, 99, 9, and 49 network evaluations, respectively.

For classwise coverage, real images use true labels and generated images are
grouped independently by frozen classifiers with seeds 17 and 29, without
confidence filtering. Coverage uses the classifiers' 128-dimensional features
and the kNN rule above. Weighting classwise coverage by real class counts
recovers overall recall.

ALD-posterior starts at $t=1-1/N$ with
$\bx_{1-1/N}=m_{1-1/N}\bar\bx_{\mathrm{data}}+
\sqrt{V_{1-1/N}}\bn$, avoiding division by $m_1=0$. This mean-based
initialization omits $m_{1-1/N}^2\operatorname{Cov}(\bx_0)$ even at the
population mean. Replacing the clean image in the adjacent Gaussian
conditional by $\hat\bx_0^{\mathrm P}$ also omits
\begin{equation}
 m_{t-\Delta t}^2
 \left(1-\frac{m_t^2V_{t-\Delta t}}
                 {m_{t-\Delta t}^2V_t}\right)^2
 \operatorname{Cov}(\bx_0\mid\bx_t)\succeq0.
 \label{eq:ald_posterior_missing_covariance}
\end{equation}
The sampler makes $N-1$ network evaluations, uses the covariance in
Eq.~\eqref{eq:ald_posterior_covariance}, and clips only the final image to
$[-1,1]$. Numerical variance and scale floors are finite-precision safeguards.

\subsection{CIFAR-10LT Augmentation Protocol}

The 100{:}1 CIFAR-10LT class counts are
$(5000,2997,1796,1077,645,387,232,139,83,50)$. All class-conditional
generators use the same CIFAR U-Net, batch size 100, learning rate
$2\times10^{-4}$, and 400{,}000 updates. DDPM and DLPM use 100 diffusion
steps, and ALD-posterior uses 100 network evaluations. Each generator produces
50{,}000 images.

A ResNet-18 teacher trained on the 50{,}000 balanced CIFAR-10 training images
assigns hard pseudo-labels without confidence filtering. Each augmentation arm
trains a new ResNet-18 from 50{,}000 real and 50{,}000 synthetic images. Every
batch has 64 real and 64 synthetic images. Training uses 10{,}000 SGD updates,
initial learning rate 0.1, momentum 0.9, weight decay $5\times10^{-4}$, and
cosine learning-rate decay. Evaluation uses the balanced 10{,}000-image test
set. Classwise F$_1$ is one-vs-rest from the multiclass confusion matrix.
}

\subsection{CelebA64 Sampling Details}
\label{app:celeba64_sampling}

{
\paragraph{Sampling configurations.}
Each method retains its own training objective and reverse sampler:
Gaussian ancestral DDPM, fixed-$\alpha$ DLPM
($\alpha=1.75$, scale cap 40) with its native posterior,
and ALD-posterior ($\alpha_{\min}=1.5$, $\tau=0.1$, $S_{\max}=20$)
with the generalized posterior update below.
DDPM and ALD-posterior use the cosine VP schedule
\citep{nichol2021improved}, with
\[
 m_t=
 \frac{\cos\!\left(\frac{t+0.008}{1.008}\frac{\pi}{2}\right)}
      {\cos\!\left(\frac{0.008}{1.008}\frac{\pi}{2}\right)},
 \qquad \sigma_t^2=1-m_t^2,
 \qquad 0\leq t\leq t_{\max}=0.9946.
\]

\paragraph{Generalized ALD-posterior update.}
For a budget of $N$ network evaluations, define the descending grid
\[
 t_j=t_{\max}\left(1-\frac{j}{N+1}\right)^p,
 \qquad j=0,\ldots,N+1.
\]
Write $m_j=m_{t_j}$ and $\sigma_j=\sigma_{t_j}$.
Starting with $V_{N+1}=0$, independently draw the truncated scales
$\tilde S_{t_j}$ using $\alpha(t_j)$ and accumulate
\[
 \gamma_j=\frac{m_j}{m_{j+1}},\qquad
 V_j=\gamma_j^2V_{j+1}
     +(\sigma_j^2-\gamma_j^2\sigma_{j+1}^2)\tilde S_{t_j},
 \qquad j=N,\ldots,1.
\]
Sampling starts at $t_1$ from
$\bx_{t_1}=m_1\bar\bx_{\mathrm{data}}+\sqrt{V_1}\,\bn$,
where $\bar\bx_{\mathrm{data}}$ is the empirical training-data mean
and $\bn\sim\mathcal N(\mathbf0,\mathbf I)$.
At each $j=1,\ldots,N$, set
\[
 s_j=\operatorname{clip}_{[10^{-6},S_{\max}]}
       (V_j/\sigma_j^2),\qquad
 \hat\eps_j=\eps_{\btheta}(\bx_{t_j},t_j,\log s_j),\qquad
 \hat\bx_{0,j}=\frac{\bx_{t_j}-\sqrt{V_j}\,\hat\eps_j}{m_j}.
\]
Following the variance parameterization used in DDIM
\citep{song2021ddim}, define
\[
 \lambda_j=1-\frac{\gamma_j^2V_{j+1}}{V_j},\qquad
 r_j^2=\eta^2\lambda_jV_{j+1},\qquad 0\leq\eta\leq1,
\]
and update
\[
 \bx_{t_{j+1}}=m_{j+1}\hat\bx_{0,j}
       +\sqrt{V_{j+1}-r_j^2}\,\hat\eps_j+r_j\bn_j,
 \qquad \bn_j\overset{\mathrm{i.i.d.}}\sim
        \mathcal N(\mathbf0,\mathbf I).
\]
At $\eta=1$, this reduces to the adjacent Gaussian posterior update;
at $\eta=0$, it adds no fresh Gaussian noise.
The sampler makes exactly $N$ network evaluations and clips only
the final image to $[-1,1]$.

\paragraph{Validation-only ALD sampler selection.}
Before formal evaluation, we registered nine ALD settings from
$\eta\in\{0,0.5,1\}$ and time-grid power $p\in\{1,1.5,2\}$.
Each setting generated 5,000 images at 1,000 NFE and was evaluated
against the independent validation split using all six metrics.
We formed the Pareto frontier by minimizing FID and KID and maximizing
precision, recall, density, and coverage; from its six nondominated
settings, we froze the one with lowest validation FID:
$(\eta,p)=(0.5,1.5)$ (validation FID 4.5266).
The other validation results and tradeoffs are retained in
Table~\ref{tab:celeba64_validation}.
No formal train-reference result was used to change this configuration.

\paragraph{Training and evaluation settings.}
We use the CelebA64 split with 162,770 training and 19,867 validation
images.
Following the released LIM preprocessing \citep{yoon2023lim},
the aligned images are cropped to $128\times128$ at top 57 and left 25,
resized to $64\times64$ by bilinear interpolation, and normalized to
$[-1,1]$; random horizontal flipping is used during training.
All three models follow the released LIM CelebA U-Net configuration
\citep{yoon2023lim}: base width 128, channel multipliers
$(1,2,2,2,4)$, two residual blocks per downsampling level,
attention at resolution $16\times16$ and in the bottleneck,
and dropout 0.1. ALD additionally receives the noise-scale input.
Training uses one seed, 600,000 optimizer updates, global batch size 256
(153.6 million image presentations), and an EMA decay of 0.9999.
The optimizer is AdamW with zero weight decay, initial learning rate
$10^{-4}$, linear learning-rate decay without warm-up, and
gradient-norm clipping at one.
We sample from each final EMA checkpoint with sampling seed 0,
generating 50,000 images at 1,000 network evaluations (NFE) per image.
The architecture, image count, training budget, EMA, generated count,
NFE, and real-image reference protocol are matched, while each method
retains its own training objective and reverse sampler.

\paragraph{Evaluation protocol.}
We extract the same 2,048-dimensional InceptionV3 features for generated
and real images using
\href{https://github.com/mseitzer/pytorch-fid}{\texttt{pytorch-fid}}.
FID \citep{heusel2017gans} is the mean and sample standard deviation across five fixed
50,000-image draws from the 162,770-image training pool, compared with
the same 50,000 generated images.
KID \citep{binkowski2018demystifying} uses the unbiased MMD-squared
estimator with the degree-three polynomial kernel and is reported as
the mean and sample standard deviation across 100 fixed
1,000/1,000 subsets.
Precision and recall \citep{kynkaanniemi2019improved} use $k=3$;
density and coverage \citep{naeem2020reliable} use $k=5$,
all computed by our implementation on the same fixed 10,000/10,000
real and generated feature subsets.
These are distributional kNN diagnostics, not attribute-classification
precision and recall.

\begin{table*}[t]
\centering

\scriptsize
\setlength{\tabcolsep}{4pt}
\caption{Validation-only ALD sampler sweep. Every candidate uses 5,000
generated images and 1,000 NFE; FID compares against all 19,867 validation
images, KID uses 100 subsets of 1,000, and PRDC uses fixed 5,000/5,000
subsets. KID entries are multiplied by $10^3$. Bold identifies the frozen
choice, not a winner on every metric. These values are not directly
comparable with the formal train-reference scores in
Table~\ref{tab:celeba64_formal}.}
\label{tab:celeba64_validation}
\begin{tabular}{lcccccccc}
\toprule
Candidate & $\eta$ & $p$ & FID$\downarrow$ & KID$\times10^3\downarrow$
& Prec.$\uparrow$ & Rec.$\uparrow$ & Dens.$\uparrow$ & Cov.$\uparrow$ \\
\midrule
p100\_e000 & 0 & 1.0 & 4.6452 & 1.2289 & 0.7302 & 0.6216 & 1.1331 & 0.9566 \\
p100\_e050 & 0.5 & 1.0 & 4.6430 & 1.3127 & 0.7384 & 0.6172 & 1.1388 & 0.9554 \\
p100\_e100 & 1 & 1.0 & 5.0132 & 1.6877 & 0.7418 & 0.6090 & 1.1601 & 0.9608 \\
p150\_e000 & 0 & 1.5 & 4.5631 & 1.0842 & 0.7338 & 0.6324 & 1.1390 & 0.9606 \\
\textbf{p150\_e050} & \textbf{0.5} & \textbf{1.5} & \textbf{4.5266} & \textbf{1.1273}
& \textbf{0.7452} & \textbf{0.6224} & \textbf{1.1823} & \textbf{0.9618} \\
p150\_e100 & 1 & 1.5 & 4.7560 & 1.2562 & 0.7612 & 0.6074 & 1.2142 & 0.9626 \\
p200\_e000 & 0 & 2.0 & 4.6157 & 1.1234 & 0.7314 & 0.6268 & 1.1451 & 0.9570 \\
p200\_e050 & 0.5 & 2.0 & 4.6724 & 1.2014 & 0.7446 & 0.6238 & 1.1931 & 0.9590 \\
p200\_e100 & 1 & 2.0 & 4.9394 & 1.3779 & 0.7728 & 0.6012 & 1.2501 & 0.9676 \\
\bottomrule
\end{tabular}
\end{table*}
}

\bibliographystyle{unsrtnat}
\bibliography{references}

\end{document}